\documentclass{article}

\usepackage{microtype}
\usepackage{graphicx}
\usepackage{svg}
\usepackage{subfigure}

\def\mydefdef{\mathrel{\vcentcolon=}}

\usepackage{booktabs} 

\usepackage{hyperref}
\usepackage{color}

\usepackage[accepted]{icml2024}
\usepackage{amssymb}
\usepackage{mathtools}
\usepackage{amsthm}
\usepackage{dsfont}
\usepackage{enumitem}

\newcommand{\std}[1]{\scriptsize{$\pm$#1}}
\newcommand{\modelname}{\textsc{TimeX++}} 

\theoremstyle{plain}

\usepackage[textsize=tiny]{todonotes}

\usepackage{amsmath,amsfonts,bm}

\usepackage{amsthm}
\newtheorem{Problem}{Problem}

\def\eqref#1{equation~\ref{#1}}

\def\1{\bm{1}}

\def\mM{{\bm{M}}}

\def\mX{{\bm{X}}}

\DeclareMathAlphabet{\mathsfit}{\encodingdefault}{\sfdefault}{m}{sl}
\SetMathAlphabet{\mathsfit}{bold}{\encodingdefault}{\sfdefault}{bx}{n}

\def\sQ{{\mathbb{Q}}}
\def\sR{{\mathbb{R}}}

\def\emM{{M}}

\icmltitlerunning{\modelname: Learning Time-Series Explanations with Information Bottleneck}

\begin{document}

\twocolumn[
\icmltitle{\modelname: Learning Time-Series Explanations with Information Bottleneck}



\icmlsetsymbol{equal}{*}

\begin{icmlauthorlist}
\icmlauthor{Zichuan Liu}{nju,msft,equal}
\icmlauthor{Tianchun Wang}{psu,equal}
\icmlauthor{Jimeng Shi}{fiu}
\icmlauthor{Xu Zheng}{fiu}
\icmlauthor{Zhuomin Chen}{fiu}
\icmlauthor{Lei Song}{msft}\\
\icmlauthor{Wenqian Dong}{fiu}
\icmlauthor{Jayantha  Obeysekera}{fiu}
\icmlauthor{Farhad  Shirani}{fiu}
\icmlauthor{Dongsheng Luo}{fiu}
\end{icmlauthorlist}

\icmlaffiliation{nju}{Nanjing University}
\icmlaffiliation{msft}{Microsoft Research Asia}
\icmlaffiliation{psu}{Pennsylvania State University}
\icmlaffiliation{fiu}{Florida International University}

\icmlcorrespondingauthor{Farhad Shirani}{fshirani@fiu.edu}
\icmlcorrespondingauthor{Dongsheng Luo}{dluo@fiu.edu}

\icmlkeywords{Machine Learning, ICML}

\vskip 0.3in
]



\printAffiliationsAndNotice{\icmlEqualContribution} 

\begin{abstract}
Explaining deep learning models operating on time series data is crucial in various applications of interest which require
interpretable and transparent insights from time series signals. 
In this work, we investigate this problem from an information theoretic perspective and show that most existing measures of explainability may suffer from trivial solutions and distributional shift issues. 
To address these issues, we introduce a simple yet practical objective function for time series explainable learning. The design of the objective function builds upon the principle of information bottleneck (IB), and modifies the IB objective function to avoid trivial solutions and distributional shift issues. 
We further present \modelname, a novel explanation framework that leverages a parametric network to produce explanation-embedded instances that are both in-distributed and label-preserving. We evaluate \modelname~on both synthetic and real-world datasets comparing its performance against leading baselines, and validate its practical efficacy through case studies in a real-world environmental application. Quantitative and qualitative evaluations show that \modelname~outperforms baselines across all datasets, demonstrating a substantial improvement in explanation quality for time series data. The source code is available at \url{https://github.com/zichuan-liu/TimeXplusplus}.
\end{abstract}

\section{Introduction}
\label{sec:intro}
Deep learning has become a cornerstone technology in analyzing time series data, prevalent in scenarios such as finance~\cite{bento2021timeshap}, healthcare~\cite{kaushik2020ai}, and environmental science~\cite{adebayo2021can}. 
Despite its successes, a critical limitation of deep learning in time series analysis is the lack of explainability, which is crucial for gaining trust and actionable insights in these sensitive and impactful domains.
For instance, in environmental science, the ability to explain deep learning predictions is vital for understanding complex ecological dynamics and making informed, impactful policy decisions~\cite{reichstein2019deep,razavi2021deep,adebayo2021can}.

\begin{figure}[t]
\centering
\includegraphics[width=0.48\textwidth]{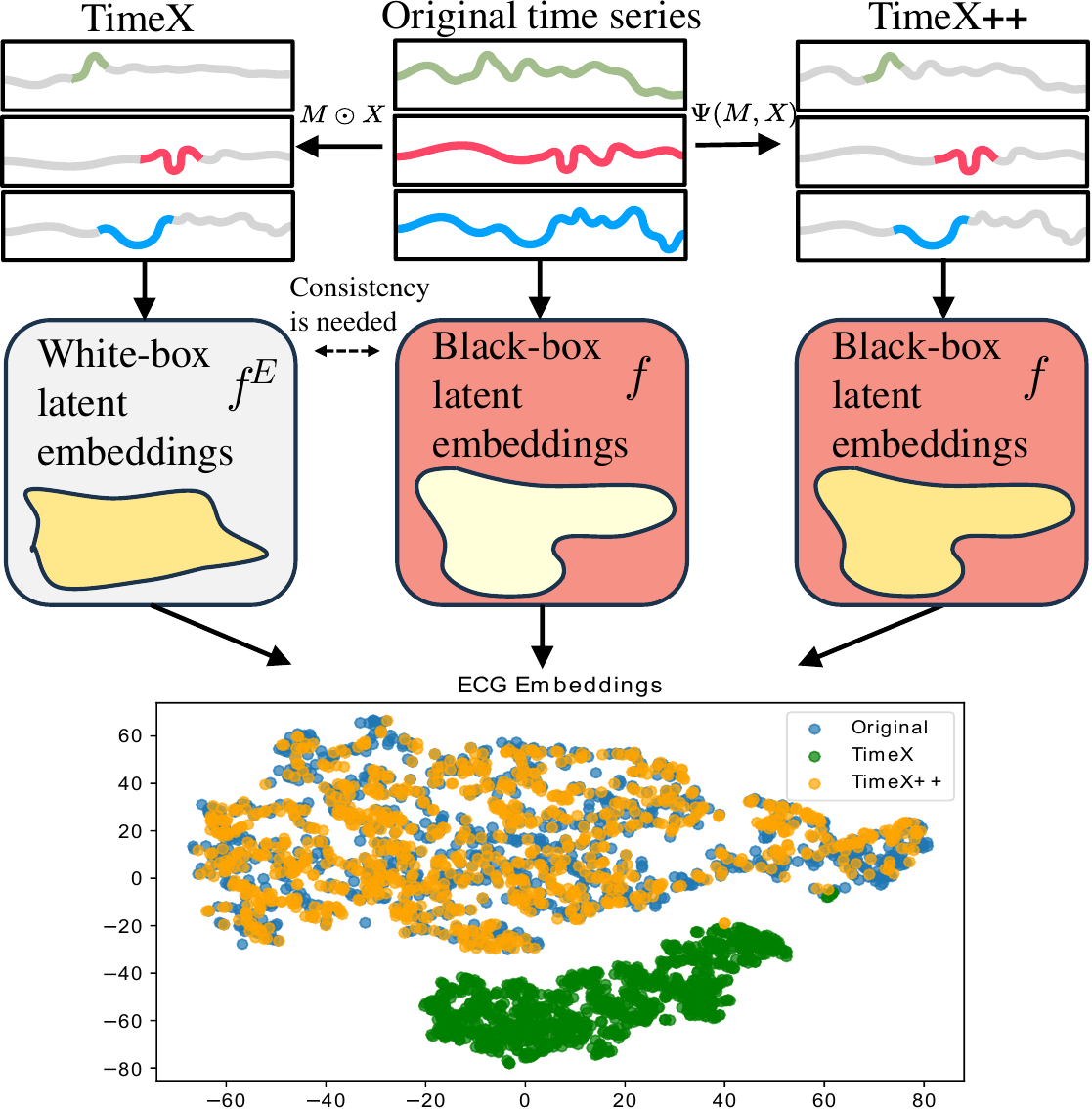}
\vspace{-3mm}
\caption{A comparison of our model and the reference model, where the latent embeddings of explanations are learned from the ECG dataset. Our explanations are within the original distribution, whereas explanations of the reference model are not.
}\label{intro}
\vspace{-4mm}
\end{figure}

Current efforts in enhancing explainability in deep learning for time series analysis primarily focus on pinpointing significant locations of time series signals that dominate the model’s prediction in a post-hoc sense. 
For example, \citet{shi2023power} explained their trained models for water level prediction using LIME, a local interpretable model-agnostic explanation technique~\cite{ribeiro2016should}.
On top of this intuitive principle, perturbation-based methods, including Dynamask~\cite{crabbe2021explaining} and Extrmask~\cite{enguehard23a}, offer insights by altering non-salient features to assess their impact on model output.
However, without a theoretical foundation, these ad-hoc designed objectives are often specific to a single domain and do not generalize to wider scenarios. 


Recently, the principle of information bottleneck~(IB) has been used for explainable learning~\cite{tishby2015deep}.  
Specifically, given a time series instance $X$ and its label $Y$, the IB principle finds an \textit{explanation} sub-instance $X'$ which  
optimizes a tradeoff between compactness and informativeness, where compactness is achieved by
minimizing the mutual information between the original instance $X$ and the sub-instance $X'$, and informativeness is achieved by maximizing the mutual information between the sub-instance $X'$ and instance label $Y$. That is, $X'$ minimizes $I(X; X')-\alpha I(X'; Y)$, where the hyperparameter $\alpha>0$ captures the trade-off between compactness and informativeness of $X'$~\cite{miao2022interpretable}. 
Thus, IB has been used extensively in domains including computer vision~\cite{luo2019significance}, natural language~\cite{mahabadi2021variational,liu2024protecting}, and graph-structured data~\cite{miao2022interpretable}.

The application of the IB principle in the context of time series explainability faces several limitations. 
First, due to the potentially large size of the time series data $X$, accurate estimation of the mutual information term $I(X; X')$ may require significant amounts of training data (e.g., \cite{mcallester2020formal}). 
Second, as shown in previous studies~\cite{huang2024factorized}, the IB loss function is sometimes minimized by sub-instances that do not align with the intuitive notion of explainability, leading to perceptually unrealistic explanations. This latter issue, which is referred to as the \textit{signaling} issue, arises when an explanation rule produces sub-instances $X'$ which \text{signal} the value of $Y$, e.g., by taking very large values when $Y=1$ and small values when $Y=0$, without necessarily aligning with the notion of explainability. Such sub-instances have high $I(X';Y)$ due to the high correlation between $X'$ and $Y$ and low $I(X;X')$ due to the fact that $X'$ only signals the value of $Y$, however, they do not provide an intuitively valid explanation despite optimizing the IB loss function. 

An alternative approach to evaluating the suitability of explanations is to apply the classifier $f(\cdot)$ directly to the sub-instance $X'$ and evaluate a cross-entropy term $\mathrm{CE}(Y; Y')$, where $Y'=f(X')$. 
This resolves the aforementioned signaling issue and is used by existing methods~\cite{enguehard23a, liu2024explaining, queen2023encoding} hereinafter called reference models. 
However, $X'$ is often out-of-distribution for $f(\cdot)$ thus leading to inaccurate predictions for $Y'$~\cite{zhao2022ood}. To elaborate, as shown in Figure~\ref{intro}, let us take \textsc{TimeX}~\cite{queen2023encoding} as an example (other reference models face a similar problem), where the distribution of original instances and their sub-instances is substantially different from each other. Then, since $f(\cdot)$ is trained on the original instances, it would produce inaccurate results when applied to the explanation sub-instances.
Consequently, \textsc{TimeX} retrains an additional white-box model $f^E(\cdot)$ for model consistency, while explanation instances (generated by the reference models) input to $f(\cdot)$ remain unreliable. 
Such replication requires knowledge of the to-be-explained model structure, and consistency in the behavior of the white-box model may not equal consistency in explanation.

To address these challenges, we develop a practical objective function for time series explainable learning. 
Specifically, we replace the compactness quantifier $I(X; X')$ in IB with a traceable variational upper-bound, which consists of a minimality constraint and a discrete constraint. 
To avoid the signaling issue, we replace the informativeness quantifier $I(X'; Y)$ with a measure of label consistency between $Y'$ and $Y$, where $Y'$ is the label of sub-instance $X'$. 
We propose a novel explainable framework \modelname~that generates in-distributed and label-preserving time series instances for both quantifiers. 
\modelname~maintains the new instances consistent with the original distribution (see Figure~\ref{intro}) and preserves the uninformative areas provided by the reference model.
We summarize our contributions as follows  
\begin{itemize}[leftmargin=*]
\item We investigate the limitations of existing explanation models for time series learning from the perspective of information theory and propose a practical objective function. 
\item We propose a novel explanation framework \modelname, which addresses the distribution shifting issue by generating in-distributed and explanation-embedded instances.
\item We achieve state-of-the-art performances compared to other explainers on eight synthetic and real-world time series datasets, and verify the effectiveness in a real-world application from environmental science.  
\end{itemize}

\section{Notations and Preliminary}

\subsection{Notations and Problem Formulation} 
This work focuses on explainability in time series classification. A time series instance $X\in \mathcal{X} = \mathbb{R}^{T\times D}$ is represented by a $T\times D$ real-valued matrix, where $T$ is the length of the time series, and $D$ is the feature dimension. A multivariate time series is one for which $D>1$, otherwise, the time series is called univariate. The value of the feature indexed $d$ at time $t$ is denoted by $X[t,d]$. A training set  $\mathcal{T}=\{(X_i,Y_i)|i\in [N]\} $ consists of $N$ time series instances $X_i$ along with their associated labels $Y_i$, where $Y_i\in \mathcal{C}$ and $\mathcal{C}=\{1,2,\cdots,|\mathcal{C}|\}$ is the set of all possible labels. We use the shorthand $\mX=\{X_i\}_{i=1}^N$ to refer to the instances without labels.  A time series classifier $f(\cdot)$ takes an instance $X\in \mathbb{R}^{T\times D}$ as input, and outputs a label $f(X)\in \mathcal{C}$.  

In order to develop a generally applicable model for explainability, we consider explanation methods which are task-agnostic and treat the to-be-explained model $f(\cdot)$ as a black box, i.e., the so-called \textit{post-hoc, instance-level} explanation methods ~\cite{zhang2021survey}. 
In this context, an explanation refers to a sub-instance of the input time series, extracted using a saliency mask, which is a `sufficient statistic' of the input with respect to its label. 
Furthermore, we consider model explainability, rather than task explainability. That is, the extracted sub-instance must be a sufficient statistic with respect to the model output, rather than the ground-truth label~\cite{faber2021comparing, liu2024explaining}. The high-level problem statement is given as follows. 
\begin{Problem} [Post-hoc Instance-level Time Series Explanation]
\label{prob:exp}
Given a trained model $f$ and input $X \in \mathcal{X} =\sR^{T\times D}$, the objective in post-hoc instance-level time series explanation is to find a sub-instance $X'$ that `explains' the prediction of $f$ on $X$.  
The sub-instance $X'$ is obtained by applying a binary mask $M \in \mathcal{M} =\{0,1\}^{T\times D}$ on $X$, i.e.,  $X' = X \odot M$, where $\odot$ is the element-wise multiplication.
\end{Problem}

As can be observed in the problem statement, in order to find good post-hoc instance-level time series explanations, given an observed instance $X$, one needs to optimize the choice of binary mask $M\in \{0,1\}^{T\times D}$ with respect to an underlying objective function, e.g., the information bottleneck objective function discussed in the subsequent sections. 
In this work, we transform this discrete optimization problem into a continuous one, and consider stochastic masks. 
That is, we define an explanation extractor $g(\cdot)$ as a function that takes the instance $X$ as input, and outputs a matrix $\bm{\pi}=[\pi_{t,d}]_{t\in [T],d\in [D]}\in [0,1]^{T\times D}$. Then, the binary mask is generated by producing each $M[t,d]$ independently and according to a Bernoulli distribution with parameter $\pi_{t,d}$.

\subsection{The Information Bottleneck Principle}
The IB principle ~\cite{tishby1999information} has been widely used in the explainability literature. 
Formally, given an input instance $X$ with label $Y$, the IB principle obtains a compact and informative sub-instance $X'$ of $X$ using the following optimization:
\begin{equation}
    \label{eq.ib}
    \max_{\substack{g: \mathcal{X}\mapsto [0,1]^{T\times D}
     \\M[t,d]\sim \mathrm{Bern}(\pi_{t,d})}} I(Y; X') -\alpha I(X; X'),
\end{equation}
where $X'=X\odot M $, $[M_{t,d}]_{t\in [T],d\in [D]}$ are generated independently and according to a Bernoulli distribution with parameter $\pi_{t,d}$, $g(X)=\bm{\pi}= [\pi_{t,d}]_{t\in [T],d\in [D]}$,
and $\alpha$ is a hyperparameter capturing the trade-off between informativeness quantified by $I(Y; X')$ and compactness quantified by $I(X; X')$. 
The IB principle has been used to obtain several explanation mechanisms ~\cite{mahabadi2021variational, wu2020graph}. However, it has been pointed out that IB-based approaches suffer from a \textit{signaling} issue which may lead to explanation outputs that do not align with the notion of explainability. That is, the value of $X'$ can \textit{signal} the value of $Y$, yielding a large $I(X;Y)$ and small $I(X;X')$, without necessarily aligning with the intuitive notion of explainability. 
The interested reader is referred to Appendix \ref{App:sig} for an illustrating example of the signaling issue.   
To avoid this issue, the IB optimization is modified as follows:
\begin{equation}
    \label{eq.ib2}
     \min_{\substack{g: \mathcal{X}\mapsto [0,1]^{T\times D}
     \\M[t,d]\sim \mathrm{Bern}(\pi_{t,d})}} -\mathrm{LC}(Y; Y')+\alpha I(X; X'),
\end{equation}
where $\mathrm{LC}(Y;Y')$ denotes the label consistency (LC) between $Y$ and  the label of $X'$, denoted by $Y'$, which is discussed in detail in the subsequent sections. It can be noted that the signaling issue, described through an example in Appendix \ref{App:sig}, is resolved by this modification.

\section{Explaining Time Series Learning via Information Bottleneck}

\begin{figure*}[t]
\centering \vspace{1mm}
\includegraphics[width=0.8\textwidth]{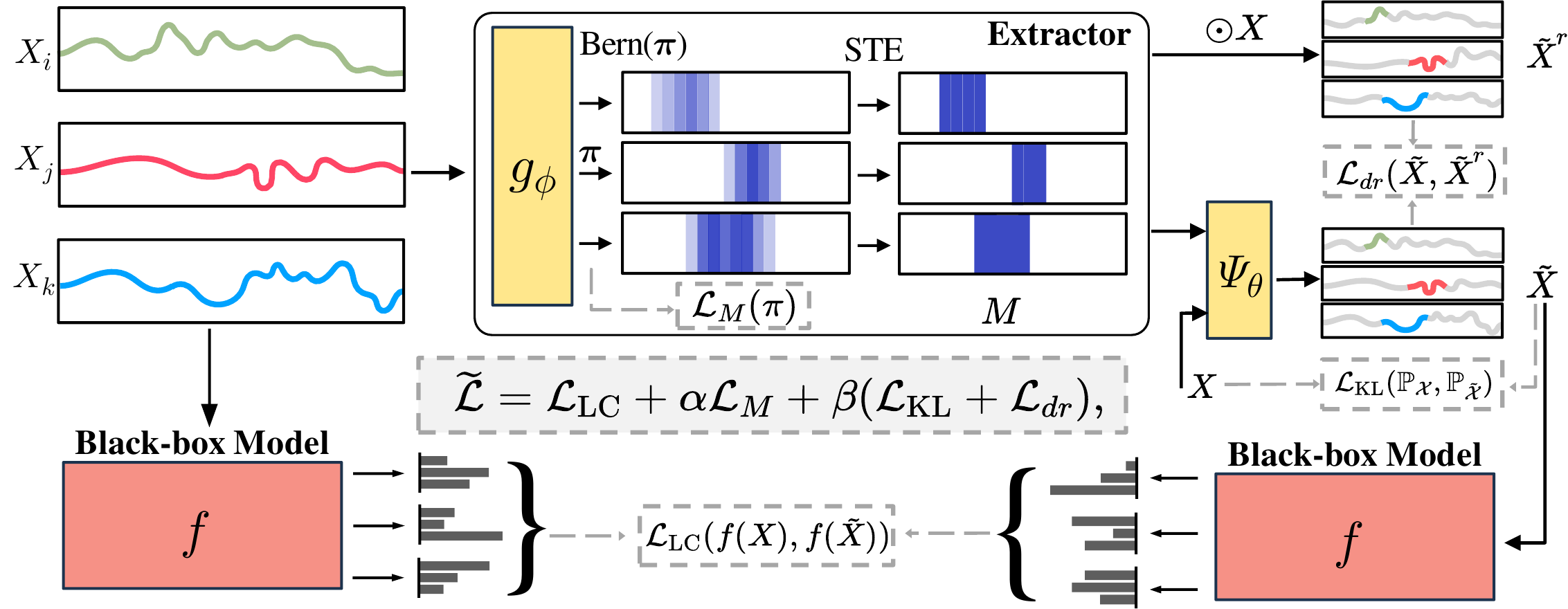}
\vspace{-1mm}
\caption{The overall architecture of \modelname, which consists of the explanation extractor and the conditioner generating new instances.
}\label{framwork}
\vspace{-1mm}
\end{figure*}

As discussed in the prequel, the application of the IB principle to explainability problems suffers from the signaling issue. 
Additionally,  direct application of the modified IB principle in Eq.~(\ref{eq.ib2}) in time series explanation problems is challenging due to the complex and high-dimensional nature of the data~\cite{goldfeld2020information}. 
Specifically, the temporal dependencies and varying feature dynamics inherent in time series complicate the accurate estimation of mutual information, a key component of the IB framework, thus rendering the direct application of IB computationally intractable in this context. 
In the following, we identify several additional issues in applying the IB principle, and derive a tractable objective to address these issues.

\textbf{Modifying the Compactness Quantifier $I(X; X')$.} 
As mentioned in the previous sections,
the term $I(X;X')$ is included in IB to ensure the compactness of the resulting explanation sub-instance. That is, to ensure that the explanation contains as few features and time-instances as needed. However, in the context of time-series classification, the mutual information term $I(X;X')$ does not necessarily align with this objective, as it sometimes gives preference to multiple low-entropy sub-instances of $X'$ as opposed to a single high-entropy sub-instance, thus violating compactness. An illustrating example of this issue is provided in Appendix \ref{App:size}.  
To address this, we modify the IB optimization further, and consider the following objective function:
\begin{equation}
    \label{eq.ib4444}
     \min_{\substack{g: \mathcal{X}\mapsto [0,1]^{T\times D}
     \\M[t,d]\sim \mathrm{Bern}(\pi_{t,d})}}\!\!\!\!\!\!\!\! -\mathrm{LC}(Y; Y')\! + \!\mathbb{E}_X[\alpha \!\!\sum_{t,d}\!\! H(M[t,d])+ \gamma|M|],
\end{equation}
where $\alpha,\gamma>0$ are hyperparameters and $H(\cdot)$ is the entropy, 
We provide the following justification for this modified objective function. 
The term $I(X;X')$ was included in the original IB formulation to ensure the compactness of the resulting sub-instance explanation. Minimizing $\mathbb{E}(|M|)$ serves a similar purpose, by limiting the average number of non-zero elements in the mask $M$, while avoiding the aforementioned problem which arises due to high-entropy components of the time series sequence in optimizing the original IB objective. 
The inclusion of the term $\sum_{t,d}H(M[t,d])$ ensures that the mask is `almost' deterministic, i.e., $\pi_{t,d}\approx 0$ or $\pi_{t,d}\approx 1$. 

The objective function in Eq.~(\ref{eq.ib4444}) can be further simplified in practice. Particularly
let us take a hyperparameter $r\in [0,1]$ and define $\sQ(\emM[t,d]) \sim \mathrm{Bern}(r)$ and $\sQ(M) = \prod_{t,d} \sQ(\emM[t,d]$). Then, as shown in Appendix \ref{App:simplify}, Eq.~(\ref{eq.ib4444})  can be reformulated as:
\begin{equation}\label{eq4}
\begin{aligned}
     \min_{\substack{g: \mathcal{X}\mapsto [0,1]^{T\times D}
     \\M[t,d]\sim \mathrm{Bern}(\pi_{t,d})}} & -\mathrm{LC}(Y; Y') \\
     &+\alpha \mathbb{E}_X \left [ D_\mathrm{KL}(\mathbb{P}({M}|X)\|\mathbb{Q}(M)) \right ],
    \end{aligned}
\end{equation}
where we have defined $\mathbb{P}({M}|X)=\prod_{t,d}\mathbb{P}({M[t,d]|X})$, and for a given instance $X$,  $\mathbb{P}({M[t,d]|X})$ is a Bernoulli distribution with parameter $\pi_{t,d}$.

\textbf{The Informativeness Quantifier $\mathrm{LC}(Y; Y')$.}
In the modified IB formulation in Eq.~(\ref{eq4}), the infromativeness of explanations is quantified by $\mathrm{LC}(Y; Y')$ which measures the cross entropy between the original label $Y$ and the label $Y'$ generated based on the explanation sub-instance $X'$. 
Ideally, one would produce $Y'$ by directly applying $f(\cdot)$ on $X'$. However, this approach suffers from the out-of-distribution (OOD) issue as described in the following. The sub-instance $X'$ is not necessarily in-distribution for $f(\cdot)$ and hence directly applying $f(\cdot)$ to $X'$ may not yield an accurate estimate of the correct label. That is, while $X'$ is a sub-instance of $X$, and it can be observed as part of $X$ in the training data, it may not be observed in isolation as an element of the training set. Thus, $f(\cdot)$ is not trained to classify $X'$ directly, and will not necessarily yield accurate predictions if applied to this input.
 
A routinely adopted method for producing $Y'$ is to `pad' the sub-instance $X'$ using additional Gaussian noise variables to form a time series instance $\widetilde{X}^r$, and then setting $Y'=f(\widetilde{X}^r)$ (e.g., ~\citet{fong2017interpretable, enguehard23a}). 
Formally, we consider an approximate baseline distribution: $\mathbb{B}_\mathcal{X}  = \Pi_{t, d} \mathcal{N} (\mu_{t,d}, \sigma^2_{t,d})$, where $\mu_{t,d}, \sigma^2_{t,d}$ are the mean and variance over the whole time series samples $\mX$.
Then, the `padded' instance  is given by:
\begin{equation}\label{bbbbbb}
    \widetilde{X}^{r}\mydefdef M\odot X + (1- M)\odot b, \text{ where } b\sim \mathbb{B}_\mathcal{X}.
\end{equation}
The justification is that while $X'$ is OOD with respect to $f(\cdot)$, it might be the case that using an appropriate choice of padded input parameters $\mu_{t,d},\sigma^2_{t,d}$, 
$\widetilde{X}^{r}$ may be made in-distribution for $f(\cdot)$.

In practice, although transforming $X'$ to $\widetilde{X}^r$ using the padding approach alleviates the OOD issue to some extent, it does not completely mitigate the problem. 
An alternative method has been introduced recently to address the OOD issue in \textsc{TimeX}~\cite{queen2023encoding}. In that approach, first, a copy of the classifier $f(\cdot)$, denoted by $f^{E}(\cdot)$ is constructed. Then,  
to address the OOD challenge, \textsc{TimeX} fine-tunes the parameters of $f^{E}(\cdot)$ with a consistent loss between $f(X)$ and $f^{E}(\widetilde{X}^r)$, so that $f^{E}$ is trained on instances of $\widetilde{X}^r$ and $f(\cdot)$ on instances of $X$. However, this state-of-the-art method suffers from two main drawbacks. 
First, \textsc{TimeX} has to treat the to-be-explained model as a white box whose architecture and model parameters are accessible (see Figure~\ref{intro}). However, in a wide range of real-world scenarios, the classifier model is given in a black-box manner. 
Second, similar to the original IB, a cross-entropy loss defined between $Y$, the ground-truth label, and $Y'$ the output of $f^{E}(\widetilde{X}^r)$ suffers from the signaling issue. That is, $\widetilde{X}^r$ can signal the value of $Y$ and $f^{E}$ may be trained to detect the signal, thus yielding explanations that do not align with the intuitive notion of explainability (please refer to the example in Appendix \ref{App:sig} which explains the signaling problem in detail). 

To tackle these limitations, we propose an additional step in generating in-distributed instances from $X'$. That is, we first generate $\widetilde{X}^r$ using the standard padding technique described previously. Then, we take $\widetilde{X}^r$ as a \textit{reference instance} and generate a new \textit{explanation-embedded instance} $\widetilde{X}\in  \widetilde{\mathcal{X}}= \mathbb{R}^{T\times D}$ by minimizing two loss functions: i) a loss function  $\mathcal{L}_{\mathrm{KL}}(\mathbb{P}_{\mathcal{X}}, \mathbb{P}_{\widetilde{\mathcal{X}}})$, quantifying the distribution shift between $\widetilde{X}$ and $X$, and ii) a loss function  $\mathcal{L}_{dr}(\widetilde{X},\widetilde{X}^{r})$, quantifying Euclidean distance between $\widetilde{X}^r$ and $\widetilde{X}$. The exact formulation of each of these loss functions is discussed in detail in the following sections. 
The first loss function ensures that $\widetilde{X}$ is in-distribution for $f(\cdot)$, while the second loss function builds upon the previously adopted methods mentioned in the prequel, and by forcing $\widetilde{X}$ to have a distribution that is `close' to a Gaussian distribution, ensures a general solution without overfitting. 
Thus, to maximize $\mathrm{LC}(Y; Y')$, the total loss function for the informativeness of the explanation is equal to:
\begin{equation}\label{eq8}
\begin{aligned}
    \mathcal{L}_\mathrm{LC}(f(X),f(\widetilde{X})) +\beta (\mathcal{L}_{\mathrm{KL}}(\mathbb{P}_{\mathcal{X}}, \mathbb{P}_{\widetilde{\mathcal{X}}})  + \mathcal{L}_{dr}(\widetilde{X},\widetilde{X}^{r})).
\end{aligned}
\end{equation}









\section{\modelname~Method}
This section describes the \modelname~ approach, which builds upon the ideas set forth in the previous sections.  
The architecture is depicted in Figure~\ref{framwork}.
Specifically, we introduce an explanation extractor $g_\phi$ to learn based on the compactness quantifier of Eq.~(\ref{eq4}). 
Next, an explanation conditioner $\Psi_\theta$ controls the generation of explanation-embedded instances, which must generate in-distribution instances $\widetilde{X}$ using the loss in Eq.~(\ref{eq8}). 
 \modelname~learns label consistency for reliable approximation of $\mathrm{LC}(Y;Y')$. Each of these components is detailed in the following. 

\textbf{Explanation Extractor $g_\phi$.} The extractor $g_\phi: \mathbb{R}^{T\times D }\mapsto  [0,1]^{T\times D}$   encodes the input $X$ into the stochastic mask $\bm{\pi}$.
We implement the 
compactness quantifier 
in Eq.~(\ref{eq4}) by parameterizing $g_\phi(\cdot)$ via an encoder-decoder transformer,  in which $\mathbb{P}(M|X)$ is represented by $g_\phi(\cdot)$.
This parameterization can be understood as a way to assign attribution scores such that a low $\pi_{t,d}$ has a low probability of being masked-in.
Furthermore, to penalize irregular non-contiguous shapes in a time series sample, \modelname~also optimizes a connective loss $\mathcal{L}_{\text{con}}$ for the predicted distributions:
\begin{equation}
\label{eq.con}
\begin{aligned}
\mathcal{L}_{\text{con}} =  \lambda_{\mathrm{con}}\frac{1}{T\times D}  \sum_{d=1}^D \sum_{t=1}^{T-1} \sqrt{\left(\pi_{t,d}-\pi_{t+1,d}\right)^2},
\end{aligned}
\end{equation}
where $\lambda_{\mathrm{con}}$ is the penalization strength.
Further, as discussed in \citet{queen2023encoding}, the explanation generating stochastic mask should be `hardened' via mapping to a deterministic binary mask.
Thus, we adopt a straight-through estimator (STE)~\cite{jang2016categorical} to obtain a binary mask $M \in \{0,1\}^{T\times D}$, i.e, $M := \text{STE}(M)$. 
As discussed in Appendix \ref{App:simplify}, the loss function of masks binding Eq.~(\ref{eq.con}) when training $g_{\phi}(\cdot)$ can be written as:
\begin{equation}
\begin{aligned}
\mathcal{L}_M= & \sum_{t,d} [\pi_{t,d}\log(\frac{\pi_{t,d}}{r}) \\
 + & (1-\pi_{t,d})\log(\frac{1-\pi_{t,d}}{1-r})] + \mathcal{L}_{\text{con}}.
\end{aligned}
\end{equation}

\textbf{Explanation Conditioner $\Psi_\theta$.} A key idea in the design of \modelname~is to use a two-step process to generate the instance $\widetilde{X}$ which in turn is used to quantify informativeness. That is, to first generate the reference instance $\widetilde{X}^r$ using the conventional Gaussian padding method described in the prequel, and then to generate $\widetilde{X}$ to mitigate the OOD issue. 
We directly $\widetilde{X}$ learn via a parameterized conditioner $\Psi_\theta$ so that:
\begin{equation}
\begin{aligned}
\widetilde{X} =  \Psi_\theta(M, X),
\end{aligned}
\end{equation}
where $\Psi_\theta$ is an MLP that maps the concatenation $[M, X]$ into $\widetilde{X}\in \widetilde{\mathcal{X}} = \mathbb{R}^{T\times D}$.
To ensure the in-distribution property, we minimize the KL divergence:
\begin{equation}\label{eq12}
\begin{aligned}
&\mathcal{L}_{\mathrm{KL}}(\mathbb{P}_\mathcal{X}, \mathbb{P}_{\tilde{\mathcal{X}}})  =\mathbb{E}\left[D_{\mathrm{KL}}\left(\mathbb{P}_\mathcal{X}(X) \| \mathbb{P}_{\tilde{\mathcal{X}}}(\tilde{X}) \right)\right].
\end{aligned}
\end{equation}
As mentioned previously, 
to ensure generalizability and avoid overfitting, we minimize the Euclidean distance between $\widetilde{X}$ and the Gaussian padded reference $\widetilde{X}^r$ using:
\begin{equation}
\begin{aligned}
\mathcal{L}_{dr}(\widetilde{X},\widetilde{X}^{r}) = \frac{1}{T\times D}\sum_{d=1}^D \sum_{t=1}^{T}||\widetilde{X}[t, d]- \widetilde{X}^{r}[t, d]||^2.
\end{aligned}
\end{equation}

\textbf{Label Consistency.} To maintain a similar amount of label in $\mathrm{LC}(Y; Y')$, we use the preservation game~\cite{fong2017interpretable} that minimizes the deviation of predictions from the original ones by the explained black-box $f(\cdot)$. We use the same Jensen-Shannon (JS) divergence as \textsc{TimeX} for the label consistency loss $\mathcal{L}_{\mathrm{LC}}$ in Eq.~(\ref{eq8}) by:
\begin{equation}
\begin{aligned}
\mathcal{L}_{\mathrm{LC}} (f(X), f(\widetilde{X})) = \mathbb{E}\left[D_{\mathrm{JS}} (f(X) \| f(\widetilde{X})) \right].
\end{aligned}
\end{equation}

\textbf{Overall Learning Objective.} The learning objective is to train the whole framework by minimizing the total loss:
\begin{equation}\label{eq13}
\begin{aligned}
\widetilde{ \mathcal{L} }= \mathcal{L}_{\mathrm{LC}} + \alpha \mathcal{L}_M + \beta(\mathcal{L}_{\mathrm{KL}} + \mathcal{L}_{dr}),
\end{aligned}
\end{equation}
where $\{ \alpha, \beta \}\in \mathbb{R}$ are hyperparameters adjusting the weight of losses.
\modelname~is optimized end-to-end, requiring little hyperparameter tuning. 
The choice of $r$ for regularized masks is still crucial and proved to be stable~\cite{miao2022interpretable}. 
Hence, we set $r = 0.5$ to remain consistent throughout experiments, which is analyzed in Appendix~\ref{ablations}.
We summarize the pseudo-code of \modelname~in Appendix~\ref{pseudo}.

\section{Experiments}\label{sec:exp}

\begin{table*}[t]
\scriptsize 
    \centering
    \caption{Attribution explanation performance on univariate and multivariate synthetic datasets.}
    \resizebox{1.8\columnwidth}{!}{
    \begin{sc}
    \begin{tabular}{l|c c c| c c c }
         \toprule
         & \multicolumn{3}{c|}{FreqShapes} & \multicolumn{3}{c}{SeqComb-UV} \\
         Method & AUPRC & AUP & AUR & AUPRC & AUP & AUR \\
         \midrule
         IG & {0.7516}\std{0.0032} & {0.6912}\std{0.0028} & {0.5975}\std{0.0020} & {0.5760}\std{0.0022} & 0.8157\std{0.0023} & {0.2868}\std{0.0023} \\
         Dynamask & 0.2201\std{0.0013} & 0.2952\std{0.0037} & 0.5037\std{0.0015} & 0.4421\std{0.0016} & {0.8782}\std{0.0039} & 0.1029\std{0.0007}\\
         WinIT & 0.5071\std{0.0021} & 0.5546\std{0.0026} & 0.4557\std{0.0016} & 0.4568\std{0.0017} & 0.7872\std{0.0027} & 0.2253\std{0.0016}\\
         CoRTX & 0.6978\std{0.0156} & 0.4938\std{0.0004} & 0.3261\std{0.0012} & 0.5643\std{0.0024} & 0.8241\std{0.0025} & 0.1749\std{0.0007}\\
         SGT + Grad & 0.5312\std{0.0019} & 0.4138\std{0.0011} & 0.3931\std{0.0015} & 0.5731\std{0.0021} & 0.7828\std{0.0013} & 0.2136\std{0.0008}\\
         TimeX & \underline{0.8324}\std{0.0034} & \underline{0.7219}\std{0.0031} & \underline{0.6381}\std{0.0022} & \underline{0.7124}\std{0.0017} & \textbf{0.9411}\std{0.0006} & \underline{0.3380}\std{0.0014} \\
          \midrule
         \modelname & \textbf{0.8905}\std{0.0018} & \textbf{0.7805}\std{0.0014} & \textbf{0.6618}\std{0.0019} & \textbf{0.8468}\std{0.0014} & \underline{0.9069}\std{0.0003} & \textbf{0.4064}\std{0.0011} \\
         \midrule
         \midrule
         & \multicolumn{3}{c|}{SeqComb-MV} & \multicolumn{3}{c}{LowVar} \\
         Method & AUPRC & AUP & AUR & AUPRC & AUP & AUR \\ 
         \midrule
         IG & 0.3298\std{0.0015} & {0.7483}\std{0.0027} & 0.2581\std{0.0028} & \underline{0.8691}\std{0.0035} & {0.4827}\std{0.0029} & {0.8165}\std{0.0016}\\
         Dynamask & 0.3136\std{0.0019} & 0.5481\std{0.0053} & 0.1953\std{0.0025} & 0.1391\std{0.0012} & 0.1640\std{0.0028} & 0.2106\std{0.0018}\\
         WinIT & 0.2809\std{0.0018} & 0.7594\std{0.0024} & 0.2077\std{0.0021} & 0.1667\std{0.0015} & 0.1140\std{0.0022} & 0.3842\std{0.0017} \\
         CoRTX & 0.3629\std{0.0021} & 0.5625\std{0.0006} & 0.3457\std{0.0017} & 0.4983\std{0.0014} & 0.3281\std{0.0027} & 0.4711\std{0.0013}\\
         SGT + Grad & 0.4893\std{0.0005} & 0.4970\std{0.0005} & \textbf{0.4289}\std{0.0018} & 0.3449\std{0.0010} & 0.2133\std{0.0029} & 0.3528\std{0.0015}\\
         TimeX & \underline{0.6878}\std{0.0021} & \underline{0.8326}\std{0.0008} & {0.3872}\std{0.0015} & {0.8673}\std{0.0033} & \underline{0.5451}\std{0.0028} & \textbf{0.9004}\std{0.0024}\\
                  \midrule
         \modelname & \textbf{0.7589}\std{0.0014} & \textbf{0.8783}\std{0.0007} & \underline{0.3906}\std{0.0011} & \textbf{0.9466}\std{0.0015} & \textbf{0.8057}\std{0.0016} & \underline{0.8332}\std{0.0016}\\
         \bottomrule
    \end{tabular}
    \end{sc}
    }
    \vspace{-4mm}
    \label{tab:attr_synthetic}
\end{table*}

\begin{table*}[t]
\scriptsize 
    \centering
    \caption{Difference between the distribution of different explanation instances and the distribution of original data.}
    \resizebox{1.8\columnwidth}{!}{
    \begin{sc}
    \begin{tabular}{l|c c c| c c c }
         \toprule
         & \multicolumn{3}{c|}{FreqShapes} & \multicolumn{3}{c}{SeqComb-UV} \\
         Method & KDE $\uparrow$& KL-Divergence $\downarrow $& MMD & KDE $\uparrow$& KL-Divergence $\downarrow $& MMD $\downarrow $\\
         \midrule
         Zero& -36.6705\std{ 0.2747} & \underline{0.1440}\std{0.0069} & 0.0769\std{0.0044} & -90.0931\std{ 0.3115} & 0.3985\std{0.0048} & 0.1668\std{0.0032} \\
         Mean &\underline{-36.5394}\std{0.1942} & 0.1465\std{0.0034} & 0.0777\std{0.0063} &\underline{-83.0381}\std{0.2708} & \underline{0.3249}\std{0.0089} & 0.1112\std{0.0047}\\
         $b\sim \mathbb{B}_\mathcal{X}$ & -53.8552\std{0.5086} & 0.2888\std{0.0032} & \underline{0.0240}\std{0.0004} &-100.0496\std{0.6577} & 0.4981\std{0.0045} & \textbf{0.0085}\std{0.0004}\\
         Ours & \textbf{-28.7566}\std{2.6582} & \textbf{0.0964}\std{0.0239} & \textbf{0.0162}\std{0.0037} & \textbf{-57.9323}\std{1.6837} & \textbf{0.0598}\std{0.0176} & \underline{0.0777}\std{0.0186}\\
         \midrule
         \midrule
         & \multicolumn{3}{c|}{SeqComb-MV} & \multicolumn{3}{c}{LowVar} \\
         Method & KDE $\uparrow$& KL-Divergence $\downarrow $ & MMD $\downarrow $& KDE $\uparrow$& KL-Divergence $\downarrow $& MMD $\downarrow $\\
         \midrule
         Zero & -257.0395\std{1.6558} & 0.7394\std{0.0290} & 0.2476\std{0.0063}& -431.9932\std{1.1306} & 0.5972\std{0.0116} & 0.1049\std{0.0029}\\
         Mean &\underline{-230.2502}\std{1.3131} & \underline{0.4549}\std{0.0192} & 0.0741\std{0.0089} &\underline{-429.3312}\std{1.2488} & \underline{0.5710}\std{0.0118} & \underline{0.0953}\std{0.0035}\\
         $b\sim \mathbb{B}_\mathcal{X}$ &-260.1821\std{1.5705} & 0.7213\std{0.0188} & \underline{0.0261}\std{0.0002} &-474.4819\std{1.3187} & 0.9763\std{0.0166} & 0.1041\std{0.0001}\\
         Ours & \textbf{-191.2647}\std{0.8897} & \textbf{0.0377}\std{0.0099} & \textbf{0.0141}\std{0.0116} & \textbf{-426.4307}\std{2.7648} & \textbf{0.5380}\std{0.0251} & \textbf{0.0725}\std{0.0172}\\
\bottomrule
    \end{tabular}
    \end{sc}
    }
    \vspace{-1mm}
    \label{tab:perturbations}
\end{table*}

\begin{table*}[t]
    \centering
    \scriptsize
    \caption{(\textit{Left}) Attribution explanation performance on the ECG dataset. (\textit{Right}) Results of ablation analysis.}
    \setlength{\tabcolsep}{3pt}
    \resizebox{1.8\columnwidth}{!}{
    \begin{sc}
    \begin{tabular}{l|c c c||l| c c c}
         \toprule
         & \multicolumn{3}{c||}{ECG} & \modelname & \multicolumn{3}{c}{ECG}\\
         Method & AUPRC & AUP & AUR & Ablations & AUPRC & AUP & AUR\\
         \midrule
         IG & {0.4182}\std{0.0014} & \underline{0.5949}\std{0.0023} & 0.3204\std{0.0012} & Full & \textbf{0.6599}\std{0.0009} & \underline{0.7260}\std{0.0010} & \underline{0.4595}\std{0.0007}\\
         Dynamask & 0.3280\std{0.0011} & 0.5249\std{0.0030} & 0.1082\std{0.0080} &  w/o STE & 0.6152\std{0.0007} &  \textbf{0.7468}\std{0.0008} & 0.4023\std{0.0012}\\
         WinIT & 0.3049\std{0.0011} & 0.4431\std{0.0026} & {0.3474}\std{0.0011} & w/o  $\mathcal{L}_{\mathrm{LC}}$ & 0.6209\std{0.0019} & 0.6417\std{0.0020} & 0.4287\std{0.0015}\\
         CoRTX & 0.3735\std{0.0008} & 0.4968\std{0.0021} & 0.3031\std{0.0009} & 
         w/o $\mathcal{L}_{\mathrm{KL}}$ & \underline{0.6417}\std{0.0019} & 0.6979\std{0.0009} & 0.4424\std{0.0007}\\
         SGT + Grad & 0.3144\std{0.0010} & 0.4241\std{0.0024} & 0.2639\std{0.0013} & w/o $\mathcal{L}_{dr}$  & 0.1516\std{0.0003} &  0.1405\std{0.0003} & \textbf{0.6313}\std{0.0006}\\
         TimeX & \underline{0.4721}\std{0.0018} & {0.5663}\std{0.0025} & \underline{0.4457}\std{0.0018} & w/o $\mathcal{L}_{\mathrm{con}}$ & 0.6072\std{0.0008}& 0.6921\std{0.0010} & 0.4387\std{0.0007}\\
        \midrule
         \modelname & \textbf{0.6599}\std{0.0009} & \textbf{0.7260}\std{0.0010} & \textbf{0.4595}\std{0.0007}\\
         \cmidrule{1-4}
    \end{tabular}
    \end{sc}
    }
    \label{tab:saliency_ecg}
\end{table*}

\begin{figure*}[t]
    \centering
    \includegraphics[width=0.98\linewidth]{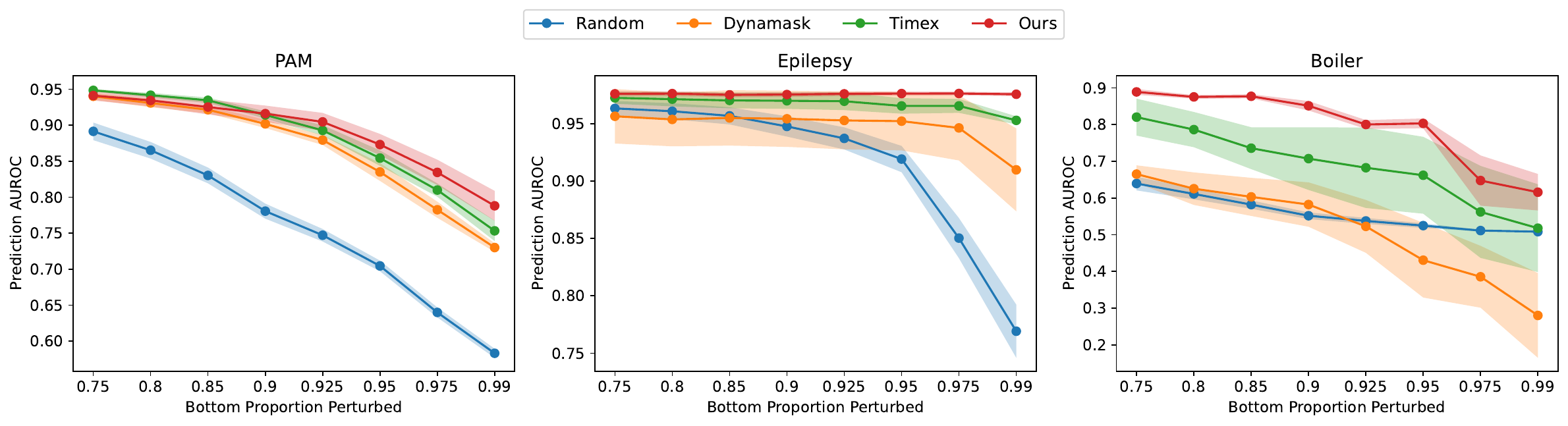}
    \vspace{-3mm}
    \caption{Occlusion experiments on real-world datasets. Higher values indicate better performance.}
    \vspace{-2mm}
    \label{fig:occlusion}
\end{figure*}

In this section, we evaluate the quality of our explanations on four synthetic datasets and six real-world datasets.
We employ a Transformer~\cite{vaswani2017attention} classifier as the black-box model $f$ to explain, where the hyperparameters are optimized to ensure model performance.
In each evaluation metric, we mark \textbf{bold} as the best and \underline{underline} as the second best.
All reported results for our method, baselines, and ablations are presented as mean $\pm$ std from $5$ fold cross-validation.
Additionally, we evaluate the feasibility of our approach by conducting a case study involving real-world flood prediction.
The details of each dataset and black-box model are provided in Appendix~\ref{datadatils}.

\subsection{Feature Importance for Synthetic Datasets}
\textbf{Datasets and Benchmarks.} We first conduct experiments on four 
datasets with known ground-truth explanations: \textbf{FreqShapes}, \textbf{SeqComb-UV}, \textbf{SeqComb-MV}, and \textbf{LowVar}.
These datasets are meticulously curated to encapsulate a wide array of temporal dynamics within both univariate and multivariate settings.
In this context, it is crucial to highlight the importance of identifying key features at different time steps.
We compare our method with six explainability benchmarks, including integrated gradients~(IG)~\cite{sundararajan2017axiomatic}, Dynamask~\cite{crabbe2021explaining}, WinIT~\cite{leung2023temporal},  CoRTX~\cite{chuang2023cortx}, SGT + GRAD~\cite{ismail2021improving}, and \textsc{TimeX}~\cite{queen2023encoding}. The implementation details of all algorithms are available in Appendix~\ref{datadatdadsils}.

\textbf{Metrics.} Given that the precise salient features are known, we utilize them as the ground truth for evaluating explanations.
At each time step, features causing prediction label changes are attributed an explanation of $1$, whereas those that do not effect such changes are $0$.
Following \citet{crabbe2021explaining}, we evaluate the quality of explanations with area under precision (AUP) and area under recall (AUR). 
We also employ AUPRC for consistency with~\citet{queen2023encoding}, which combines the results from both AUP and AUR. 
For all evaluated metrics higher values are better, and their detailed definitions are referred to the Appendix~\ref{metrics}.

\textbf{Results.} Table~\ref{tab:attr_synthetic} summarizes the performance results of the above explainers on univariate and multivariate datasets.  
\modelname~outperforms other explainers on $9$ out of $12$ cases (across $3$ metrics on four datasets), achieving an average improvement in explanation AUPRC of $11.01\%$, AUP of $10.87\%$, and AUR of $1.25\%$ when compared to the strongest baseline \textsc{TimeX}.
Note that under all these datasets, AUP is more valuable than AUR because the predicted signal has redundant information.
We analyze the statistical significance of multiple methods using the Friedman Test, resulting in the statistic $F_F=51.32$ and $p<0.001$ for all methods on $12$ cases. 
This suggests that there are significant differences in the results among the different methods.
Specifically, when considering the global metric AUPRC, \modelname~significantly improves ground-truth explanation $6.97\%$ on FreqShapes, $18.86\%$ on SeqComb-UV, $10.33\%$ on SeqComb-MV, and $8.91\%$ on LowVar over strongest baselines.
It matches our claim that our method provides explanation instances without hurting the predictions trained via the IB principle.
We also provide visualizations of salient subsequences in Appendix~\ref{visme}.

\textbf{Distribution Analysis of Perturbations.} 
To verify that the explanation instances $\widetilde{X}$ and $\widetilde{X}^r$ are within the distribution of original datasets, we utilize kernel density estimation\footnote{\url{https://scikit-learn.org/stable/modules/generated/sklearn.neighbors.KernelDensity}} (KDE) to assess the log-likelihood of each explanation instance under the original distribution, approximating zero indicates a higher likelihood of explanation instances originating from the original distribution.
We also quantify the KL divergence and maximum mean discrepancy~(MMD) between the distribution of original instances and explanation instances, where a smaller value means a greater similarity between the two distributions.
For $\widetilde{X}^r$, we pick three baselines in Eq.~(\ref{bbbbbb}): including $b$ is Zero, $b$ is Mean of all $\mX$, and $b\sim \mathbb{B}_\mathcal{X}$.
We perform experiments on synthetic datasets and provide the results in Table~\ref{tab:perturbations}.
The findings reveal that the explanation instance by our \modelname~more closely resembles the original distribution.
It significantly reduces the probability of OOD relative to the strongest baseline, which finding is consistent with the visualization depicted in Figure~\ref{intro}.
It indicates that our reliable approximation creates in-distributed and explanation-embedded instances.
We also present the distribution of explanation embeddings over several datasets, which are shown in Appendix~\ref{embemb}.

\subsection{Feature Importance for Real-world Datasets}
\textbf{Datasets and Benchmarks.} We similarly evaluate our method on four datasets from real-world time series classification tasks: \textbf{ECG}, \textbf{PAM}, \textbf{Epilepsy}, and \textbf{Boiler}.
Extracting wave intervals by \citet{queen2023encoding}, the ground-truth explanations of ECG are defined as the QRS interval, where arrhythmias can be detected.
Thus it can be used as a real-world evaluation, while other datasets without ground-truth explanations use occlusion experiments~\cite{tonekaboni2020went, crabbe2021explaining, liu2024explaining}.
Besides, we further expand the occlusion experiments with two datasets \textbf{Water} and \textbf{Freezer}, which are typical time series data in the UCR archive~\citep{dau2019ucr}.
For benchmarks and metrics, we maintain consistency with the synthetic experiments previously employed.

\begin{table*}[t]
\scriptsize
\begin{center}
\caption{Performance report on different real-world datasets by masking the top 10\% of salient features. The masked portion is substituted with an average of this feature or with zeros.}  
\label{top_feature}
\centering
\resizebox{2\columnwidth}{!}{
    \begin{sc}
\begin{tabular}{ll|ccccc|c}
\toprule
 \multicolumn{2}{l|}{Method \quad Substitution}& PAM & Epilepsy & Boiler& Wafer & Freezer & Rank\\
\midrule
 Random & Mean & 	0.9791\std{0.0018} & 0.9394\std{0.0045} & 0.8142\std{0.0555}& 0.9925\std{0.0023} & 0.7716\std{0.0932} & 7.2\\
  & Zero & 0.9794\std{0.0018} & 0.9395\std{0.0043} & 0.8950\std{0.0120}& 0.9920\std{0.0021} & 0.7747\std{0.0930} & 7.8\\
 Dynamask & Mean & 0.7776\std{0.0154} & 0.8155\std{0.0194} & 0.8080\std{0.0364} & 0.4878\std{0.2672} & 0.4515\std{0.1288} & 4.6\\
  & Zero & 	0.7765\std{0.0154} & \underline{0.3612}\std{0.0725} & 0.5553\std{0.1459	} & 0.4882\std{0.2671} & \textbf{0.3550}\std{0.1100} & 3.6\\
 TimeX& Mean & \underline{0.7494}\std{0.0486} & 0.8518\std{0.0183} & 0.5630\std{0.0413} & 0.4633\std{0.1452} & 0.4693\std{0.0605} & 4.4\\
  & Zero & 0.7543\std{0.0493} & 0.4544\std{0.0771} & 0.4163\std{0.0476} & 0.4646\std{0.0574} & 0.4646\std{0.0574} & 3.6\\
\modelname& Mean  & \textbf{0.7172}\std{0.0194} & 0.8451\std{0.0291} & \underline{0.3851}\std{0.1229} & \underline{0.4095}\std{0.1502} & 0.3786\std{0.0738}&\underline{2.8}\\
& Zero & 0.7751\std{0.0232} & \textbf{0.3553}\std{0.1459} & \textbf{0.3612}\std{0.0725} & \textbf{0.3998}\std{0.0613} &\underline{0.3771}\std{0.0740}&\textbf{2.0}\\
\bottomrule
\end{tabular}
    \end{sc}
} 
    \vspace{-2mm}
\end{center}
\end{table*}

\textbf{Results and Ablation Study on ECG Data.}\label{abstudy}
The performance results on ECG  arrhythmia detection are presented in Tabel~\ref{tab:saliency_ecg} left.   
We can see that \modelname~outperforms the leading baselines by $39.78\%$ AUPRC, $22.04\%$ AUP, and $3.10\%$ AUR, suggesting that its performance in finding relevant QRS intervals drives the arrhythmia diagnosis.
We further explore ablation studies of our method in Table~\ref{tab:saliency_ecg} right, where w/o means no related components.
First, we show that the STE improves $7.27\%$ in AUPRC and $14.22\%$ in AUR, demonstrating the effectiveness of the STE.
We also select the label consistency loss $\mathcal{L}_{\mathrm{LC}}$, the maintenance loss $\mathcal{L}_{\mathrm{KL}}$, the loss of the reference distance $\mathcal{L}_{dr}$, and the connective loss $\mathcal{L}_{\mathrm{con}}$.
Among them, $\mathcal{L}_{dr}$ failed to assess the explanation, resulting in low AUPRC and AUP.
Moreover, the lack of other loss functions likewise produces poorer explanations compared to the base model.
Our \modelname~produces high-quality explanations, showing the value in including more intermediate states for optimizing the overall objective.
For detailed ablation experiments and choice of parameters on other datasets, see Appendix~\ref{ablations}.

\textbf{Occlusion Experiments on Real-world Datasets.}
Due to the absence of ground-truth explanations on real-world datasets, we occlude the bottom $k$-percentile of salient features to measure the change in prediction AUROC, as is done in \citet{queen2023encoding}.
Besides the above baselines, we also include a random explainer reference to control for potential misinterpretations.
The explanation results by occluding salient features are presented in Figure~\ref{fig:occlusion}.
Our results show that \modelname~outperforms others across both univariate (Epilepsy) and multivariate (PAM and Boiler) time series.
Specifically, \modelname~maintains non-decreasing performance on Epilepsy due to the retention of only salient features, and outperforms baselines at any threshold $k$ on PAM and Boiler datasets.
Moreover, our method maintains excellent stability compared to the strongest baseline \textsc{TimeX}, where the error bars of our method are narrower than it.
This is because \modelname~avoids the label leakage caused by re-optimizing a white-box predictive model.
We also delete the top 10\% of salient features and substitute them with an average of this feature or with zero perturbations~\citep{liu2024explaining}, and further expand our experiments to five real-world time series datasets. 
The results and average ranks of conducted experiments are shown in Table~\ref{top_feature}.
As can be seen, the proposed method consistently outperforms existing explainers under different perturbations. 
Among them, substitution with zeros may be more applicable to the classification of black-boxes in these datasets, and therefore it has the highest average rank. 
We have similarly analyzed the significance of multiple methods using the Friedman Test, which yields that the statistic $F_F=25.80$ and $p<0.001$ for all methods on two substitutions. 
This shows that there is a significant difference between different methods, while \modelname~performs the best.
Overall, our approach generates effective explanations through information theory.

\begin{figure}[t]	
	\centerline
	{
			\centering    
			\includegraphics[scale=0.46]{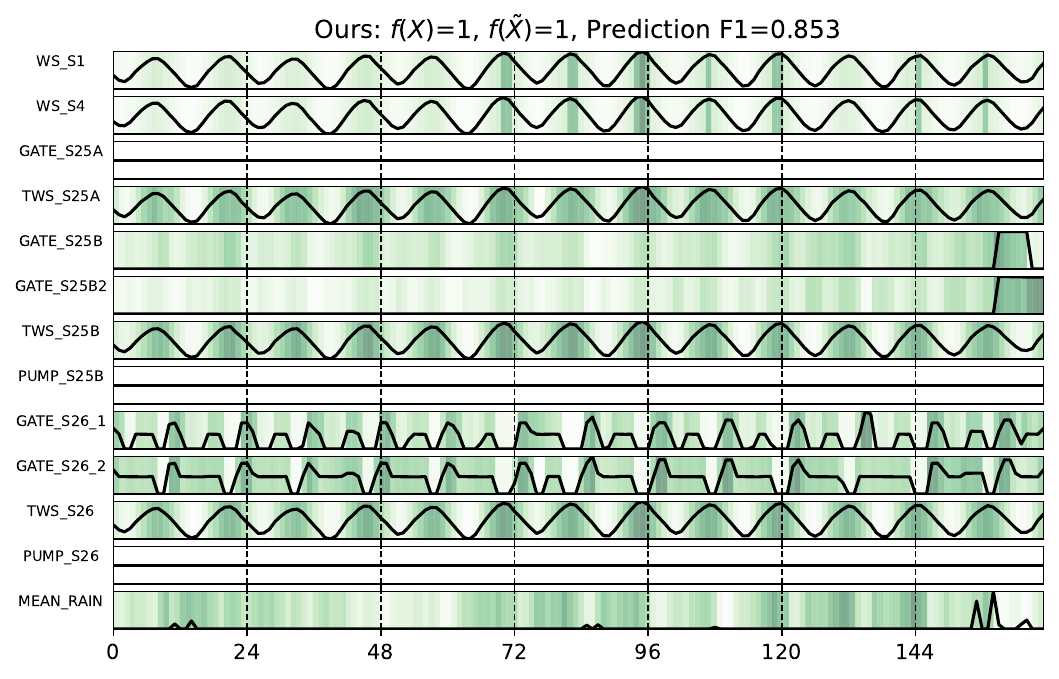}
	}
	\centerline 
	{
			\centering    
			\includegraphics[scale=0.46]{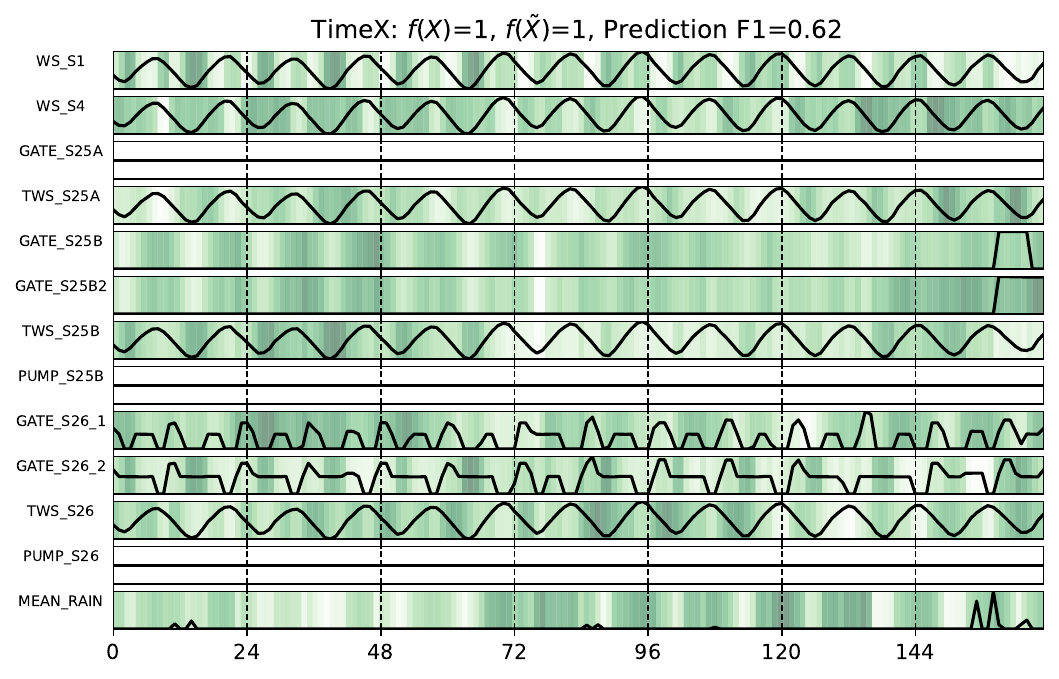}
	}\vspace{-5mm}
	\caption{A visual comparison of the explainability of our method with the leading baseline on the Florida water dataset.}\vspace{-4mm}
	\label{fig:water_explain} 
\end{figure}

\subsection{Case Study}
\textbf{Florida Water Dataset.} To assess the efficacy of our approach, we conduct a real-world case study focusing on flood prediction in a Florida coastal system~\cite{shi2023deep}. 
The detailed descriptions and a schematic diagram of the study area are illustrated in Appendix~\ref{sec:data_describe}. 
We choose to concentrate on one specific station (i.e., S1) to predict the probability of flooding in order to simplify the problem.
We use a transformer-based predictor as a black-box model.
Concurrently, we compare our model's explainability against the strongest baseline \textsc{TimeX}, specifically examining feature importance at each time step.

\textbf{Results.} 
As shown in Figure~\ref{fig:water_explain},  the first summary worth pointing out is that the F1 performance of our method outperforms that of the baseline, where the F1 metric on the top figure represents the performance of two models' explanation-embedded instance $\widetilde{X}$ in the black box. 
We break down our analysis of model explainability as follows: (i) Our \modelname~demonstrates the ability to accurately identify critical time points associated with elevated water levels, which often precede flooding. This is evident in the significance of features such as WS\_S1, WS\_S4, TWS\_S25A, TWS\_S25B, and TWS\_S26. 
(ii) When it comes to hydraulic structures like gates and pumps situated along the river's course, which control water flow from upstream to the focal point S1, our model identifies the opening of these structures as a factor that heightens the risk of flooding at S1. 
Notably, \modelname~places greater emphasis on instances of higher openings while paying less attention to lower openings. This observation can be seen in features starting with `GATE'.
(iii) Additionally, our model effectively monitors rainfall events, recognizing their substantial influence on water levels within the river.
In the context of the analysis above, the baseline \textsc{TimeX}, exhibits sporadic performance, occasionally failing to capture significant time points or focusing on unnecessary ones.

\section{Conclusion}
In this work, we theoretically investigate an information-theoretic guided objective for learning time series explanations that ensure the compactness and informativeness of explained sub-instances.
We propose a novel approach \modelname~based on the IB principle, which allows a traceability computation to produce in-distributed and label-preserving time series explanations.
Comparative studies on synthetic and real-world datasets have confirmed that
\modelname~surpasses existing explanatory tools in performance, with further proven practicality in environmental applications like flood prediction. 
Its effectiveness shows that \modelname's capability to reflect complex behaviors of pre-trained time series classifiers accurately.
However, generating sub-instances may involve some hyperparameters in the learning objective to control the quantifiers of the explanation, especially when dealing with different datasets.
We minimally tuned the hyperparameters to obtain comparable results.
Hence, it will be interesting to explore the salient areas by adopting a parameter-efficient tuning strategy.

\section*{Acknowledgements}
This project was partially supported by NSF grants IIS-2331908 and CCF-2241057. The views and conclusions contained in this paper are those of the authors and should not be interpreted as representing any funding agencies.

\section*{Impact Statement}
This paper presents work whose goal is to advance the field of Machine Learning. There are many potential societal consequences of our work, none of which we feel must be specifically highlighted here.

\bibliography{reference}
\bibliographystyle{icml2024}
\appendix
\onecolumn

\section{Related Works}

\paragraph{Explainable Artificial Intelligence.} 
With the ongoing advancement of neural networks, there is a burgeoning necessity to facilitate user comprehension of their operational mechanisms~\cite{danilevsky2020survey}. 
Predominantly, the corpus of research within explainable artificial intelligence (XAI) has been concentrated on the disciplines of computer vision~\cite{linardatos2020explainable, buhrmester2021analysis} and natural language processing~\cite{danilevsky2020survey, jacovi2020towards}.
With their success, recent studies have been extended to explain various data modalities, including reinforcement learning~\cite{liu2023na2q}, graphs~\cite{luo2020parameterized,miao2022interpretable,luo2024towards}, and time series~\cite{bento2021timeshap, enguehard23a}.
The literature primarily focuses on \textit{in-hoc} models~\cite{agarwal2021neural, silva2020optimization} that learn intrinsic interpretability with some white-box models; and \textit{post-hoc} explainability~\cite{unified, sundararajan2017axiomatic} where explanations are provided for a trained model.
Despite these efforts, the adoption of methods for time series analysis remains relatively limited as the importance and interrelation of features shift across multivariate sequences.

\paragraph{Information Bottleneck.} IB~\cite{tishby1999information, tishby2015deep}, originally proposed for signal processing, has recently been adapted in different areas as it reduces length while preserving maximum information.
On this foundation, \citet{alemi2016deep} proposed a variational information bottleneck, which for the first time bridges the gap between deep learning and IB.
Thus, it has been wildly employed in downstream applications like semantic segmentation~\cite{luo2019significance}, summarization~\cite{mahabadi2021variational, west2019bottlesum}, defense against jailbreaks~\cite{liu2024protecting}, and graph learning~\cite{miao2022interpretable,huang2024factorized}.
However, the IB principle is less researched on multivariate time series due to the intractability of mutual information.

\paragraph{Time-series Explainability.} Recent literature~\cite{enguehard23a, bento2021timeshap} has probed into the domain of XAI with respect to multivariate time series.
Within this scope, attention-based approaches~\cite{lin2020preserving, choi2016retain} utilize attention mechanisms to generate significance scores that are inherently linked to the coefficients within the model. 
Gradient-based techniques~\cite{shrikumar2017learning, sundararajan2017axiomatic} have employed the influence of localized modifications in input on the salience of features.
Furthermore, perturbation-based methods, which are notably prevalent in time series analysis, typically modify the data via a baseline~\citep{suresh2017clinical}, generative models~\citep{tonekaboni2020went, leung2023temporal}, or by diminishing the informativeness of the data~\citep{crabbe2021explaining,liu2024explaining}.
Moreover, counterfactual-based explanations~\citep{theissler2022explainable,guidotti2022counterfactual} present minimal changes that would lead to a different classification by the model.
Despite these advances, none of these methods have explored the problem of data shifting in generating unlabeled sub-instances.
Different from them, our model generates in-distributed and label-preserving time series instances. 

\section{Example Illustrating the Signaling Problem in Applying the IB Princple}
\label{App:sig}
Let us consider a binary classification problem on a univariate time-series, where $X_i$ are independent random variables taking values from $\{-1,1\}$ with equal probability, and let $Y=\mathds{1}(X_n>0)$ be the indicator of the event that the $n$-th value in the time-series is positive, for some fixed $n\in \mathbb{N}$. Clearly, an intuitively `good' explainer must output $X_n$ as the explanation. However, this is not necessarily the output of the IB optimization as described in the following. Let us consider an explainer that outputs the maximum value of the instance $X$ if $Y=0$ and the minimum value if $Y=1$, that is:
\begin{align}
    M[t,d]
    =
    \begin{cases}
   1 &\text{if }  (t,d)=   \mathop{\arg\max}\limits_{t',d'}X[t',d'] \text{ and } Y=0\\
   1  &\text{if }  (t,d)=   \mathop{\arg\min}\limits_{t',d'}X[t',d'] \text{ and } Y=1\\
  0      &\text{otherwise}
    \end{cases}.
    \label{eq:wrong_EXP}
\end{align}
 That is, the explainer `signals' the value of $Y$ by outputting large values (equal to $1$ in this case) for $X'$ if $Y=0$ and small values (equal to $-1$ in this case), otherwise. Then, $I(X;X')\approx I(X;Y) =1$, and this choice of explainer is an optimal solution for the IB optimization as $T\to \infty$. However, the explainer does not align with the intuitive notion of a `good' explanation. 
The phenomenon was investigated in detail in~\cite{huang2024factorized} in the context of the explainability of graph neural networks. 

\section{Example Illustrating the Compactness Problem in Applying the IB Principle}
\label{App:size}
Let us consider a univariate time-series classification scenario, where for some fixed even-valued integer $n$, we have:
\begin{align*}
    X_i=
    \begin{cases}
        U_i,\qquad &\text{ if } i<n\\
        U_1+U_2+\cdots+U_{i}+N_i & \text{ if } i\geq n
    \end{cases},
\end{align*} 
where $U_i$ are independent binary symmetric random variables, i.e. $P(U_i=1)=P(U_i=0)=\frac{1}{2}$, and $N_i$ are independent Gaussian, zero-mean, unit-variance random variables. 
Let $Y= \mathds{1}(X_n>\frac{n}{2})$ be the indicator of the event that $X_n$ has a value greater than $\frac{n}{2}$. Note that $P(Y=0)=P(Y=1)=\frac{1}{2}$. Clearly, a `good' informative and compact explainer outputs $X'=X_n$ as the explanation, since $Y$ is a function of $X_n$ and is completely determined by its value. However, $I(X;X_n)=\infty$ since $X_n$ is a continuous random variable and a function of $X$. So, the IB optimization yields $X_1,X_2,\cdots,X_{n-1}$ as the explanation rather than $X_n$, which is not a compact explanation.  
Generally, if a time instance $X[n,1:D]$ is highly informative, but has high entropy, it may not be chosen by the IB principle, since it yields large $I(X;X')$, although it is a compact and informative explanation.

\section{Simplifying The Objective Function in Equation (\ref{eq.ib4444})}
\label{App:simplify}
First, let us note that by linearity of expectation, we have:
\begin{align*}
    \mathbb{E}_X\left [ |M| \right ]= \sum_{t,d} P({M}[t,d]=1)= \sum_{t,d}\mathbb{E}_X\left [\pi_{t,d}\right ].
\end{align*}
In our implementation, we set the value of $\frac{1}{T\times D} \mathbb{E}_X(|M|)= p$, and choose $p$ as a hyperparameter, which controls the sparsity of the explanation. Consequently, the objective function becomes: 
\begin{equation}
    \label{eq.ib3}
     \min_{\substack{g: \mathbb{R}^{T\times D}\mapsto [0,1]^{T\times D}
     \\M[t,d]\sim \mathrm{Bern}(\pi_{t,d})}}\!\!\!\!\! -\mathrm{LC}(Y; Y')+\alpha \mathbb{E}_X\sum_{t,d}\!\! H(M[t,d])+ \gamma TDp.
\end{equation}
Let us define $r\in [0,1]$ such that:
\[
r\triangleq 
\frac{1-\sqrt{1-2^{-\frac{\gamma p}{\alpha}+2}}}{2},
\] 
so that:
\[
\frac{1}{\alpha}\gamma p= -\log{r}-\log{1-r},\]
where we have taken $\gamma $ such that $\frac{\gamma  p}{\alpha}\geq 2$.
Then,
\begin{equation}\label{klmask}
\begin{aligned}
 &  \mathbb{E}_X [ \alpha \sum_{t,d}H(M[t,d])+ \gamma |M|]
 \\& =\mathbb{E}_X\left [ \alpha\sum_{t,d}\left( \pi_{t,d}\log{\frac{\pi_{t,d}}{r}}+
 (1-\pi_{t,d})\log{\frac{1-\pi_{t,d}}{1-r}}\right )\right ].
\end{aligned}
\end{equation}
Note that this resembles a KL-divergence term. To see this, let us define $\mathbb{Q}(M)$ as the Bernoulli distribution with parameter $r$, and $\mathbb{P}({M}|X)=\prod_{t,d} \mathbb{P}({M}[t,d]|X)$. Then, the modified objective function can be written as:
\begin{equation}
     \min_{\substack{g: \mathbb{R}^{T\times D}\mapsto [0,1]^{T\times D}
     \\M[t,d]\sim \mathrm{Bern}(\pi_{t,d})}}\!\!\! -\mathrm{LC}(Y; Y')+\alpha \mathbb{E}_X \left [ D_\mathrm{KL}(\mathbb{P}({M}|X)\|\mathbb{Q}(M)) \right ].
\end{equation}

\section{Pseudo Code}\label{pseudo} 
\begin{algorithm}[H]
	\caption{The pseudo-code of \modelname}
	\begin{algorithmic}
	\STATE	\textbf{Input:} A time series dataset $ \mathcal{T}=\{(X_i,Y_i)|i\in [N]\}$, a trained black-box predictor $f:\mathcal{X}\mapsto \mathcal{C}$, adjusting hyperparameters $\{\alpha, \beta, r, \lambda_{\mathrm{con}}\}$, total training epochs $E$, learning rate $\eta$
  \STATE	 \textbf{Output:} Mask $\mM=\left \{M_i \right \}_{i=1}^N \in \mathcal{M}$ to explain
  \STATE	 \textbf{Training:}
	\STATE Initialize a baseline distribution $\mathbb{B}_\mathcal{X}  = \Pi_{t, d} \mathcal{N} (\mu_{t, d}, \sigma^2_{t, d})$, where $\mu_{t, d}, \sigma^2_{t, d}$ are the mean and variance over $ \mathcal{T}$
 \STATE Initialize an explanation generator $g_{\phi}:\mathcal{X}\mapsto [0,1]^{T\times D}$, an  explanation conditioner $\Psi_\theta:\{\mathcal{M}, \mathcal{X}\}\mapsto \widetilde{\mathcal{X}}$
	\FOR{$e \gets 1$ to $E$}
		\FOR{$i \gets 1$ to $N$}
		\STATE Get $\bm{\pi} = g_{\phi}(X_i)$ and sample a mask $M_i \sim \mathbb{P}\left(M_i \mid X_i\right)=\prod_{t, d} \operatorname{Bern}\left(\pi_{t, d}\right)$ 
        \STATE Use a straight-through estimator $\operatorname{STE}$ to obtain the discrete mask $M_i \gets \operatorname{STE}(M_i)$
		\STATE Compute the reference instance $\widetilde{X}_i^r = M_i\odot X_i + (1- M_i)\odot b$, where $b\sim \mathbb{B}_\mathcal{X}$
		\STATE  Compute the  explanation-embedded instance $\widetilde{X}_i =  \Psi_\theta(M_i, X_i)$
  		\STATE Reparameterize the original distribution $\mathbb{P}_\mathcal{X}(X_i)$ and the explanation-embedded distribution $\mathbb{P}_{\widetilde{\mathcal{X}}}(\widetilde{X}_i)$
            \STATE Get the output predictions $f(X_i)$ and $f(\widetilde{X}_i)$, respectively
		\ENDFOR
            \STATE Regularize $\bm{\pi}$ via $\mathcal{L}_M = \mathbb{E}\left[D_{\mathrm{KL}}\left(\mathbb{P}_{\phi}\left(M \mid X\right) \| \mathbb{Q}\left( M \right)\right)\right] + \lambda_{\mathrm{con}}\frac{1}{T\times D}  \sum_{d=1}^D \sum_{t=1}^{T-1} \sqrt{\left(\pi_{t,d}-\pi_{t+1,d}\right)^2}$
            \STATE Providing uninformative areas through  $\mathcal{L}_{dr}(\widetilde{X},\widetilde{X}^{r}) = \frac{1}{T\times D}\sum_{d=1}^D \sum_{t=1}^{T}||\widetilde{X}[t, d]- \widetilde{X}^{r}[t, d]||^2$
            \STATE Make embedded instances maintain the original distribution via $\mathcal{L}_{\mathrm{KL}}(\mathbb{P}_\mathcal{X}, \mathbb{P}_{\widetilde{\mathcal{X}}}) = \mathbb{E}\left[D_{\mathrm{KL}}\left(\mathbb{P}_\mathcal{X}(X) \| \mathbb{P}_{\widetilde{\mathcal{X}}}(\widetilde{X}) \right)\right]$
            \STATE Make label consistency between the logits of both predictors through $\mathcal{L}_{\mathrm{LC}} (X, \widetilde{X}) = \mathbb{E}\left[D_{\mathrm{JS}} (f(X) \| f(\widetilde{X})) \right]$
  		\STATE Construct the total loss function as $\widetilde{ \mathcal{L} }= \mathcal{L}_{\mathrm{LC}} + \alpha \mathcal{L}_M + \beta(\mathcal{L}_{\mathrm{KL}} + \mathcal{L}_{dr})$
  		\STATE Update $\phi \gets \phi  - \eta \nabla_{\phi } \widetilde{ \mathcal{L} }$ and $\theta \gets\theta - \eta \nabla_{\theta} \widetilde{ \mathcal{L} }$
		\ENDFOR
        \STATE	 \textbf{Inference:} 
        \FOR{$i \gets 1$ to $N$}
              \STATE  Get  $\bm{\pi} = g_\phi(X_i)$ and sample final masks $M_i \sim \mathbb{P}\left(M_i \mid X_i\right) $
        \ENDFOR
        \STATE	 \textbf{Return:} Mask $\left \{M_i \right \}_{i=1}^N$
	\end{algorithmic}  
\end{algorithm}

\section{Experimental Details}\label{datadatils}

\subsection{Description of Datasets}
\label{sec:data_describe}

Following~\citet{queen2023encoding}, we conduct experiments and empirical analyses utilizing both synthetic and real-world datasets containing a total of eleven different data.
Then we delineate the constitution of each dataset category, inclusive of the methodology for the derivation of ground-truth explanations where relevant.

\paragraph{Synthetic Datasets.}
Our research incorporates the use of four synthetic datasets, which are crafted with inherent ground-truth explanations. 
These datasets include \textbf{FreqShapes}, \textbf{SeqComb-UV}, \textbf{SeqComb-MV}, and \textbf{LowVar}, respectively, and are generated in the same way described in \citet{queen2023encoding}.
The creation of these datasets adheres to established methodologies that guard against the susceptibility to heuristic learning, as articulated by~\citet{geirhos2020shortcut} and further elaborated by~\cite{faber2021comparing} within the context of graph structures. 
Each time series dataset is established using a non-autoregressive moving average (NARMA) model to introduce noise to suit the generation of synthetic time series data. 
For each synthetic dataset, we have generated 5,000 training samples, 1,000 test samples, and 100 validation samples. The description of each dataset is systematically summarized in Table \ref{tab:dataset}.

\begin{table}[t]
\centering
\caption{The description of synthetic and real-world datasets. }
    \vspace{1mm}
\label{tab:dataset}
      \resizebox{0.8\columnwidth}{!}{
    \begin{sc}
\begin{tabular}{c|ccccc}
\toprule
\textbf{Dataset} &  \textbf{\# of Samples} & \textbf{Length} & \textbf{Dimension} & \textbf{Classes} & \textbf{Task}\\ \midrule
FreqShapes & 6,100 & 50 & 1 & 4& Multi-classification\\
SeqComb-UV & 6,100 & 200 & 1 & 4&Multi-classification\\
SeqComb-MV & 6,100 & 200 & 4 & 4& Multi-classification\\
LowVar & 6,100 & 200 & 2 & 4& Multi-classification\\
\midrule
ECG & 92,511 & 360 & 1 & 5 & ECG classification\\
PAM &  5,333 & 600 & 17 & 8 & Action recognition \\
Epilepsy & 11,500 & 178 & 1 & 2 & EEG classification \\
Boiler & 160,719 & 36 & 20 & 2 & Mechanical fault detection \\
Wafer & 7,164 & 152 & 1 & 2 & Sensor classification\\
FreezerRegular & 3,000 & 301 & 1 & 2 & Sensor classification\\
\midrule
Water & 573 & 168 & 13 & 2 & Binary Classification \\
\bottomrule
\end{tabular}
    \end{sc}
    }
\end{table}

\paragraph{Real-world Datasets.}
We first employ four datasets derived from real-world time series classification tasks: \textbf{ECG}~\cite{Moody2001TheIO} for the detection of electrocardiogram arrhythmias; \textbf{PAM}~\cite{reiss2012introducing} for the recognition of human activities;  \textbf{Epilepsy}~\cite{andrzejak2001indications} for the identification of electroencephalogram seizure episodes; and \textbf{Boiler}~\footnote{\url{https://dx.doi.org/10.21227/awav-bn36}} for the automated identification of mechanical faults. These datasets are described in detail by~\citet{queen2023encoding} and the process of building them is provided, which can be accessed through this url\footnote{\url{https://doi.org/10.7910/DVN/B0DEQJ}}.
Note that only ECG has true explanatory labels, the other three datasets are only available for occlusion experiments.
We also select two representative datasets \textbf{Wafer} and \textbf{FreezerRegular} in the UCR archive~\citep{dau2019ucr} to conduct occlusion experiments, which are commonly used time series classification to explore the effects of models. 

\begin{figure*}[t]
    \centering
    \includegraphics[width=0.6\linewidth]{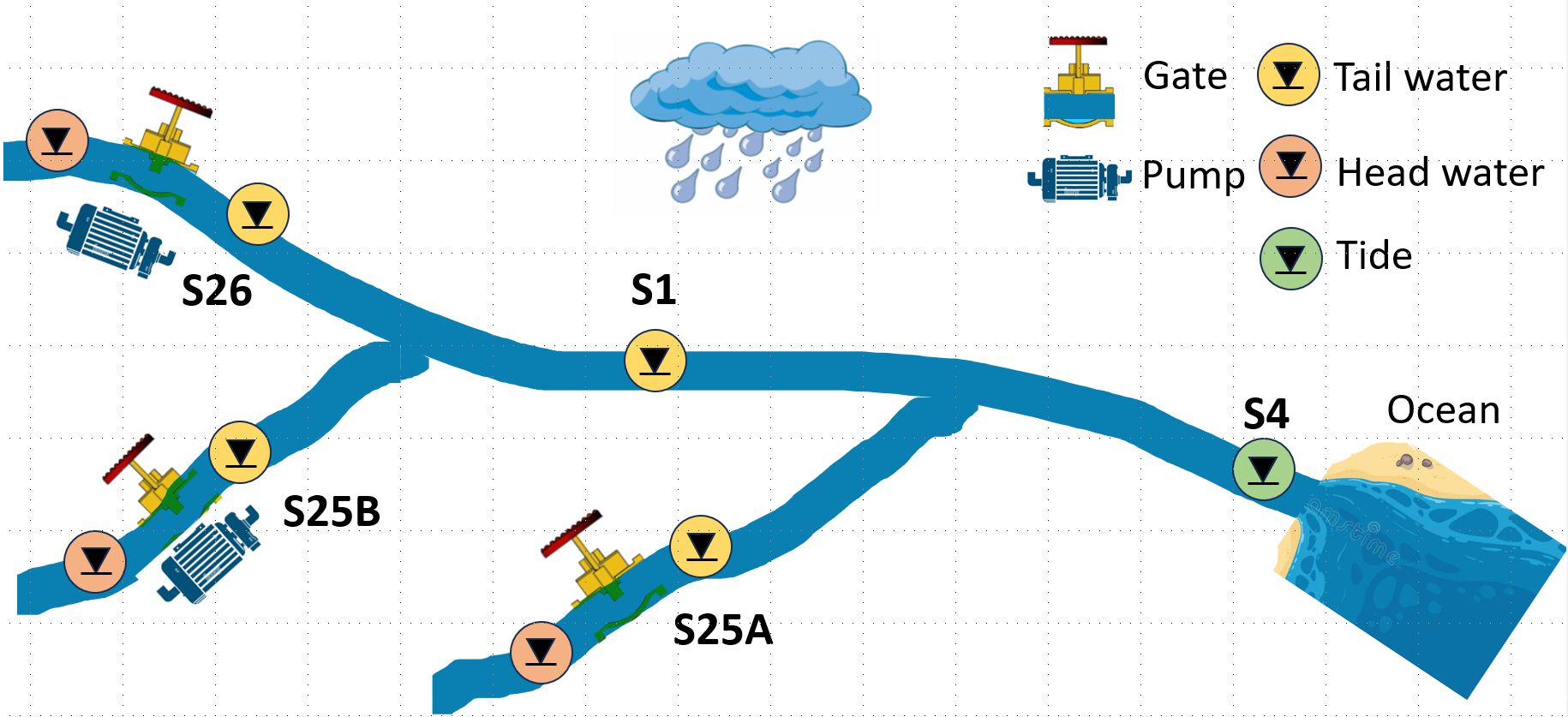}
    \caption{Diagram of the downstream of Miami River~\cite{shi2023deep}.}
    \vspace{-3mm}
    \label{fig:icml_water}
\end{figure*}

\paragraph{Florida Water Data.}  
We also conduct a real-world case study focusing on flood prediction in a South
Florida coastal system. Following \citet{shi2023deep}, we downloaded the pertinent data, including water level measurements from multiple stations, control schedules for various hydraulic structures (such as gates and pumps) along the river, tide information, and rainfall data. This dataset spans $11$ years, ranging from $2010$ to $2020$, with hourly measurements. 
We start with constructing sliding windows, each spanning one week or $168$ hours. 
These windows are tagged with labels based on the water levels observed at station S1.
A label of $1$ is assigned to a sliding window if, at any time within that window, water levels exceed the $95$-th percentile, indicating a flood warning or the presence of flooding. 
Conversely, a label of $0$ is assigned if no such conditions are met, signifying the absence of flooding.
The primary feature description and the schematic diagram of the study domain are presented in Figure~\ref{fig:icml_water}. 
We select $13$ important features for water level alarm classification, as shown in Table~\ref{mimic_features}.
Thus each sample length is $168$ and contains a $13$-dimensional feature space.
Further details can be referred to \citet{shi2023deep}.

\begin{table}[t]
    \centering
    \caption{Feature description of Florida water dataset.}
    \vspace{1mm}
      \resizebox{0.8\columnwidth}{!}{
    \begin{sc}
    \begin{tabular}{ll}
    \toprule
    Features       & Name \\
    \midrule
    Water level    & WS\_S1, WS\_S4, TWS\_S25A, TWS\_S25B, TWS\_S26 \\
    Structure  & GATE\_S25A, GATE\_S25B, GATE\_S25B2, PUMP\_S25B, GATE\_S26\_1,GATE\_S26\_2, PUMP\_S26\\
    Precipitation  & MEAN\_RAIN \\
    \bottomrule
    \end{tabular}
        \end{sc}
    }
    \label{mimic_features}
\end{table}

\subsection{Black-box Hyperparameters and Performance}
For black-box in all datasets, we employ a vanilla Transformer~\cite{vaswani2017attention} as the classification black-box model. 
\citet{faber2021comparing} suggests that A good classification performance is necessary for explainability evaluation.
Thus, we pick different hyperparameters for those black-box models as shown in Table~\ref{predictor_params_gt}, which are the same as in~\citet{queen2023encoding} to ensure the best performance and fairness with baselines.
The results of the classification performance can be found in Table~\ref{tab:clf_perf} in all datasets. 
All results guarantee the best performance as suggested by the original authors to ensure that the black-box models are strong predictors.

\begin{table}[h]
\scriptsize
    \caption{Training parameters for transformer-based predictors across all ground-truth and real-world datasets.}
    \centering
      \resizebox{1\columnwidth}{!}{
    \begin{sc}
    \begin{tabular}{c|c|c|c|c|c|c|c|c|c|c|c}
       \toprule
        Parameter & FreqShape & SeqComb-UV & SeqComb-MV & LowVar & ECG & PAM & Epilepsy  & Boiler& Wafer  & FreezerRegular& Water\\
        \midrule
        Learning rate & 0.001 & 0.001 & 0.0005 & 0.001 & 0.002 & 0.001 & 0.0001  & 0.001&0.0001 & 0.0001& 0.002\\
        Weight decay & 0.1 & 0.01 & 0.001 & 0.01 & 0.001 & 0.01 & 0.001  & 0.001&0.001 &0.001 & 0.001\\
        Epochs & 100 & 200 & 1000 & 120 & 500  & 100 & 300 & 500& 200& 300& 500\\
        \midrule
        Num. layers & 1 & 2 & 2 & 1 & 1 & 1 & 1 & 1& 1 & 1 & 1\\
        $d_h$ & 16 & 64 & 128 & 32 & 64  & 72 & 16 & 32& 16 & 16 & 64\\
        Dropout & 0.1 & 0.25 & 0.25 & 0.25 & 0.1 & 0.25 & 0.1  & 0.25& 0.1 &0.1 & 0.1\\
        Norm. embedding & No & No & No & Yes & Yes & No & No & Yes&No &No & Yes\\
        \bottomrule
    \end{tabular}
    \end{sc}
    }
    \label{predictor_params_gt}
\end{table}

\begin{table}[h]
    \scriptsize
    \centering
    \caption{The performance of transformer-based predictors for time series classification. Throughout the experimental analyses conducted in this study, these models are consistently treated as time series black-boxes.}  
      \resizebox{0.55\columnwidth}{!}{
    \begin{sc}
    \begin{tabular}{l| c c c}
         \toprule
         Dataset & F1 & AUPRC & AUROC \\
         \midrule
         FreqShapes & 0.9920\std{0.0026} & 0.9995\std{0.0003} & 0.9998\std{0.0001} \\ 
         SeqComb-UV & 0.9434\std{0.0104} & 0.9817\std{0.0061} & 0.9936\std{0.0021} \\
         SeqComb-MV & 0.9745\std{0.0063} & 0.9951\std{0.0013} & 0.9983\std{0.0005} \\
         LowVar & 0.9718\std{0.0033} & 0.9975\std{0.0006} & 0.9991\std{0.0002} \\
          \midrule
         ECG & 0.9072\std{0.0228} & 0.9345\std{0.0247} & 0.9509\std{0.0232} \\
         PAM & 0.8925\std{0.0073} & 0.9294\std{0.0042} & 0.9797\std{0.0015} \\
         Epilepsy & 0.9190\std{0.0040} & 0.9220\std{0.0074} & 0.9368\std{0.0064} \\
         Boiler & 0.8396\std{0.0107} & 0.8129\std{0.0176} & 0.8982\std{0.0132} \\
         Wafer & 0.9928\std{0.0012} & 0.9970\std{0.0009} & 0.9995\std{0.0001} \\
         FreezerRegular & 0.9793\std{0.0156} & 0.9888\std{0.0074} & 0.9902\std{0.0070} \\         
          \midrule
         Water & 0.9121\std{0.0104} & 0.9490\std{0.0125} & 0.9469\std{0.0130} \\
         \bottomrule
    \end{tabular}
    \end{sc}
    }
    \label{tab:clf_perf}
\end{table}

\subsection{Details of Our Metrics}\label{metrics}
Following \citet{crabbe2021explaining}, we use AUP and AUR as metrics to evaluate the efficacy of salient features, framing it as a binary classification task.
For an explainer, we have an obtained mask $M \in [0,1]^{T \times D}$ as its explanation.
Let $Q \in \{0,1\}^{T \times D}$ be a ground-truth matrix whose elements indicate the true saliency of the inputs contained in $X \in \mathbb{R}^{T \times D}$, where $Q[t,d] = 1$ if the feature $X[t, d]$ is salient, otherwise it is $0$. 
Let $\tau \in (0,1)$ be the detection threshold for $M[t,d]$ to indicate that the feature $X[t,d]$ is salient. 
This allows to convert the mask into an estimator $\hat{{Q}}[t,d](\tau)$ by:
\begin{align*}
\hat{Q}[t,d](\tau) = \left\{
	\begin{array}{ll}
	1 & \text{ if } M[t,d] \geq \tau \\
	0 & \text{ else.}
	\end{array}
\right.
\end{align*}
Consider truly salient index sets and index sets selected by the saliency method:
\begin{align*}
 A = & \left\{ (t,d) \in [1:T] \times [1:D] \mid Q[t,d] = 1 \right\}, \\
 \hat{A} ( \tau ) = & \left\{ (t,d) \in [1:T] \times [1:D] \mid \hat{Q}[t,d] (\tau) = 1 \right\}.
\end{align*}
The precision and recall curves that map each threshold to a precision and recall score as:
\begin{align*}
\text{P}: (0,1)  \longrightarrow  [0,1] : &  \tau  \longmapsto  \frac{\vert A \cap \hat{A}(\tau) \vert}{\vert \hat{A}(\tau) \vert}, \\
\text{R}: (0,1)  \longrightarrow  [0,1] : &  \tau  \longmapsto  \frac{\vert A \cap \hat{A}(\tau) \vert}{\vert A \vert}.          
\end{align*} 
Thus, the AUP and AUR scores can be derived by:
\begin{align*}
\text{AUP} = &  \int_{0}^1 \text{P}(\tau) d\tau, \\
 \text{AUR} = & \int_{0}^1 \text{R}(\tau) d\tau.
\end{align*}


\section{Implementation Details}\label{datadatdadsils}

\subsection{Details of Baseline Methods}
We compare our \modelname~against six popular baselines. Integrated gradients~(IG)~\cite{sundararajan2017axiomatic} is used as a general explainer; Dynamask~\cite{crabbe2021explaining} and WinIT~\cite{leung2023temporal} are used as recent time-series specific explainers; CoRTX~\cite{chuang2023cortx} is used for applied contrastive learning; SGT + GRAD~\cite{ismail2021improving} is demonstrated as an \textit{in-hoc} explainer for time series; and \textsc{TimeX}~\cite{queen2023encoding} is the strongest baseline explained by keeping the model behavior consistent. 
All hyperparameters follow the code provided by their authors, and the implementation of baselines is based on open source codes Dynamask\footnote{\url{https://github.com/JonathanCrabbe/Dynamask}} and \textsc{TimeX}\footnote{\url{https://github.com/mims-harvard/TimeX}}.

\subsection{Details of Our Method}
We opt for minimal tuning of hyperparameters, ensuring compatibility with parameters similar to those of \textsc{TimeX}.
When sparser explanations are required, the value of $\alpha$ needs to be increased to ensure the explanation extractor has compactness, $\beta$  is used to control the generation of instances within the distribution, and $\lambda$ regulates the continuity of explanations that is a smoothing constraint.
For the above nine datasets, we list hyperparameters for each experiment performed in Table~\ref{tab:timex_params_gt}.
For the explanation extractor $g_\phi$, we build a transformer encoder-decoder structure which is the same as \textsc{TimeX}, where an autoregressive transformer decoder with a $32$-dimensional feed-forward, one attention head, and a $\mathrm{sigmoid}$ activation to out probabilities for each instance to generate $\bm{\pi}$.
In the explanation conditioner $\Psi_\theta: \left \{ \mathcal{X}, \mathcal{M} \right \} \mapsto \widetilde{\mathcal{X}}$, we use a single-layer MLP with an $\mathrm{ELU}$ activation to embedd the concatenation $[M,X]$, where $M\in \mathcal{M}, X\in \mathcal{X}$ and the number of hidden layers of MLP is $32$.
See our code for more details: \url{https://github.com/zichuan-liu/TimeXplusplus}.

\begin{table}
\scriptsize
    \centering
    \caption{Training parameters for \modelname~across all ground-truth and real-world experiments.}
    \vspace{1mm}
    \resizebox{1\columnwidth}{!}{
    \begin{sc}
    \begin{tabular}{c|c|c|c|c|c|c|c|c|c|c|c}
       \toprule
        Parameter & FreqShape & SeqComb-UV & SeqComb-MV & LowVar & ECG & PAM & Epilepsy &  Boiler&  Wafer&  FreezerRegular & Water\\
        \midrule
        Learning rate & 0.001 & 0.001 & 0.002 & 0.005 & 0.0005 & 0.0005 & 0.0005  & 0.0001 & 0.0001 & 0.0005 & 0.001\\
        Batch size & 64 & 64 & 64 & 64 & 16 & 32 & 32 & 32& 64 & 64 &64\\
        Weight decay & 0.001 & 0.001 & 0.001 & 0.0001 & 0.0001 & 0.001 & 0.001 & 0.001& 0.001& 0.001 &0.01\\
        Scheduler? & Yes & Yes & No & No & No & No & Yes  & Yes&Yes &Yes &Yes\\
        Epochs & 50 & 50 & 100 & 100 & 5 & 100 & 50  & 50&100 &50 &100\\
        \midrule
        $r$ & 0.5 & 0.5 & 0.5 & 0.5 & 0.5 & 0.1 & 0.5  & 0.5&0.5 &0.5 &0.5\\
        $\alpha$ & 2.0 & 2.0 & 2.0 & 2.0 & 2.0 & 2.0 & 2.0 & 2.0&2.0 &2.0 &2.0\\
        $\beta$ & 1.0 & 1.0 & 1.0 & 1.0 & 1.0 & 1.0 & 1.0 & 1.0 & 1.0& 1.0&1.0\\
        $\lambda_{\mathrm{con}}$ & 1.0 & 2.0 & 2.0 & 2.0 & 2.0 & 0.0 & 2.0  & 2.0&  2.0&   2.0&2.0\\
        \bottomrule
    \end{tabular}
    \end{sc}
    }
    \label{tab:timex_params_gt}
\end{table}

\section{Further Distribution Analysis}\label{embemb}

\begin{figure}[t]	
	\centerline 
	{
			\centering    
			\includegraphics[scale=0.325]{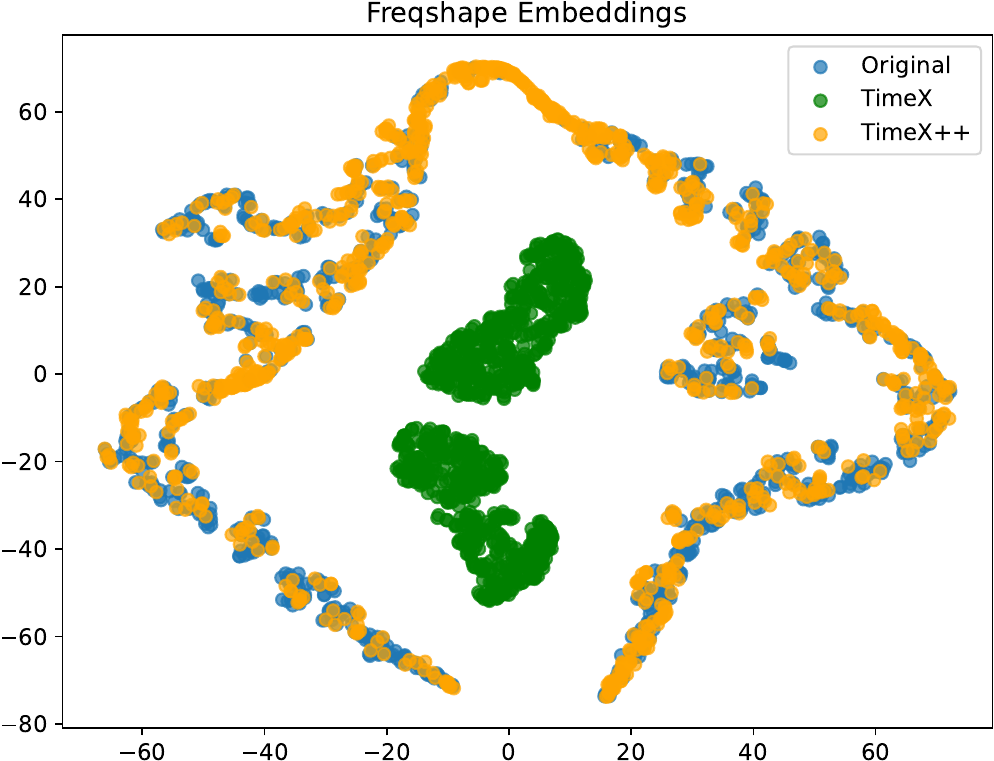}
			\includegraphics[scale=0.325]{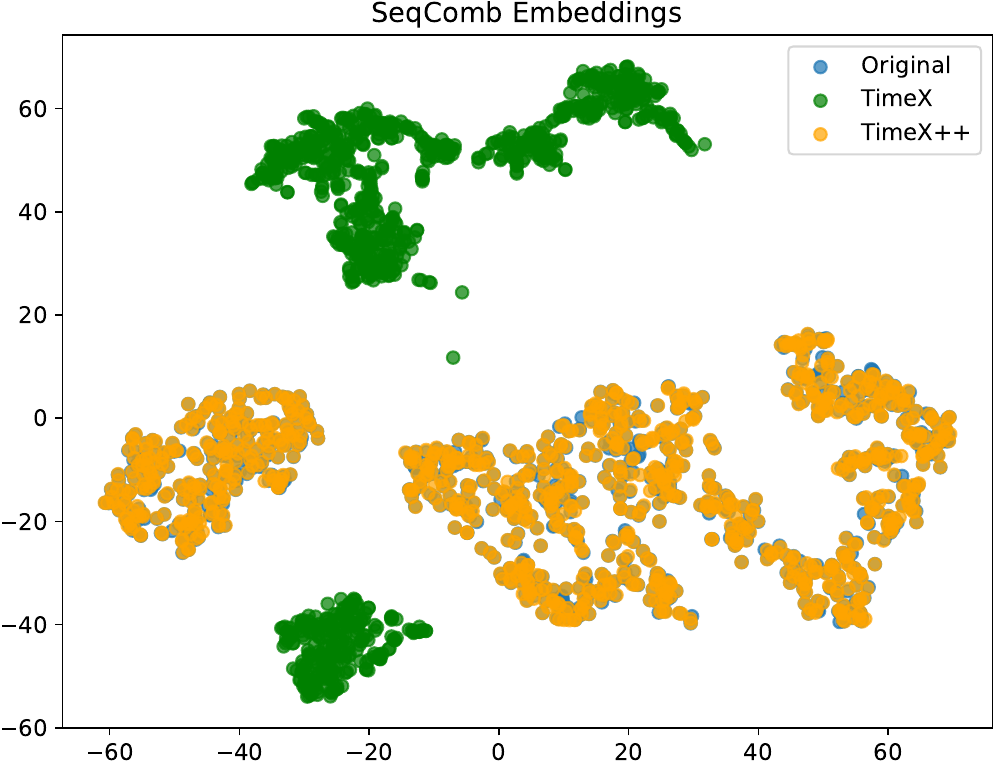}
			\includegraphics[scale=0.325]{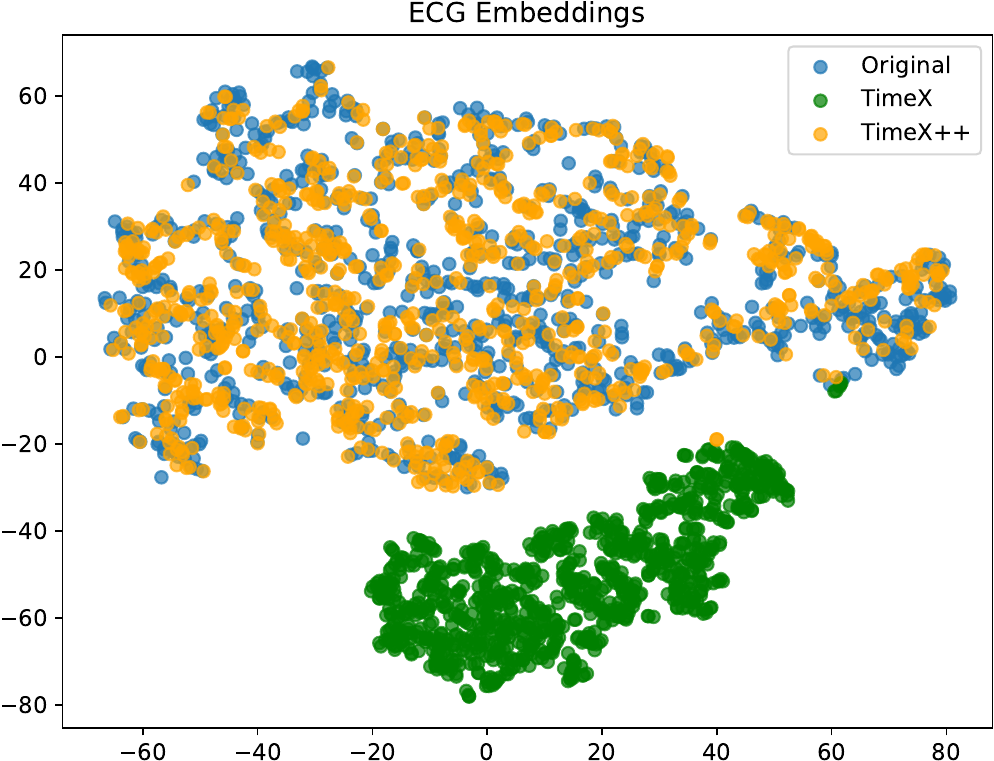}
	}
	\caption{The distribution of explanations over Freqshape (univariate), SeqComb-MV (multivariate), and ECG (real-world) datasets.}
	\label{1dasdas} 
\end{figure}

To visualize the OOD problem, we analyze the distribution between the explanation-embedded instances and the original instances produced by the different methods on three datasets: Freqshape (univariate), SeqComb-MV (multivariate), and ECG (real-world).
We plot the shape of the dataset $\mX$, where we make TSNE reduction of the time and feature level dimensions to $2$ dimensions, as shown in Figure~\ref{1dasdas}.
It is clear that there are distributional biases in the reference instances (e.g. based on \textsc{TimeX}),  whereas the distributions in our generated explanation-embedded instances by \modelname~remain consistent with those between the original instances.

\section{Computing resources and Runtime}
For computational resources, our experiments are performed on an NVIDIA 80GB Tesla A100 GPU. On average, the training runtime for each experiment in this study approximated 3-15 minutes per fold. The training time depends on the size of the data volume, e.g. the ECG dataset is large and requires a longer training time. Compared to \textsc{TimeX}, which requires training both a white-box model for model consistency and the explanation masks, we directly perturb the black-box to train in less time. We also conduct an inference runtime experiment to compare the performance of \modelname~with that of baseline explainers. We select all real-world datasets of varying sizes for this comparison. Table~\ref{times} presents the inference time of the test data in five folds, which shows the time in seconds. In all datasets, \modelname~emerges as the most expedient model during the inference phase. Such an outcome aligns with expectations given that both Dynamask and IG necessitate recursive operations for individual samples during inference, in contrast to \modelname, which requires merely a single forward propagation. For the comparison with \textsc{TimeX}, our method inference is slightly faster. This is since \textsc{TimeX} forward passes through a trained white-box model $f^E(\cdot)$ that requires Landmark calculations. Whereas \modelname~is perturbing directly on the original black-box model. Since the masks of generators have the same structure and both use transformers, the difference in training/inference time between the two is not very significant. In summary, \modelname~enables efficient inference within a second response.

\begin{table}[H]
\scriptsize
\begin{center}
\caption{Inference runtime of occlusion experiments for IG, Dynamask, \textsc{TimeX}, and \modelname~on all real-world datasets.}\label{times}
\centering
\vspace{1mm}
\resizebox{1\columnwidth}{!}{
    \begin{sc}
\begin{tabular}{l|ccccc}
\toprule
 Method & PAM & Epilepsy & Boiler& Wafer & FreezerRegular \\
\midrule
 IG & 5.2314\std{0.2474} & 15.4665\std{0.3634} & 123.5558\std{0.8070}& 15.3950\std{0.3047} & 6.7452\std{0.5017} \\
 Dynamask & 105.1664\std{0.6095} & 371.092\std{4.2567} & 3780.4326\std{468.1442} & 403.3997\std{31.6637} & 189.2849\std{1.7308}\\
 TimeX & \underline{0.1638}\std{0.2193} & \underline{0.2042}\std{0.2072} & \underline{0.9176}\std{0.2029} & \underline{0.1634}\std{0.0222} & \textbf{0.2690}\std{0.4109}\\
\modelname & \textbf{0.1611}\std{0.2179} & \textbf{0.2009}\std{0.2077} & \textbf{0.9096}\std{0.1767} & \textbf{0.1262}\std{0.0021} & \underline{0.2703}\std{0.4169}\\
\bottomrule
\end{tabular}
    \end{sc}
} 
\end{center}
\end{table}

\section{Further Ablation Experiments}\label{ablations}
In this section, we conduct a comprehensive analysis of ablation studies on the \modelname~model across three datasets: including FreqShape (univariate), SeqComb-MV (multivariate), and ECG (real-world).
This investigation serves to augment the preliminary ablation studies previously detailed for the ECG dataset in Section~\ref{abstudy}.

\paragraph{Effect of  STE.} We first conduct an experiment examining the effectiveness of using the STE for training \modelname. 
The results of our method with/without STE are shown in Table~\ref{tab:ablation_ste}.
The use of STE provided an average of $8.878\%$ increase in explanation performance compared to no STE for all datasets in AUPRC.
Moreover, the results indicate that the STE consistently enhances AUR for every dataset, while the AUP is only better for ECG without STE than with STE.
In summation, the adoption of STE unequivocally demonstrates a substantial improvement in comparison to the continuous masking strategy, thereby providing tangible support for hard masks.

\begin{table}[t]
\scriptsize 
    \centering
    \begin{center}
    \caption{Ablation of \modelname~with STE versus without STE, where if no STE denotes generating a continuous mask.}\label{tab:ablation_ste}
    \vspace{1mm}
    \resizebox{0.9\columnwidth}{!}{
    \begin{sc}
    \begin{tabular}{l|c c c| c c c }
         \toprule
         & \multicolumn{3}{c|}{Without STE} & \multicolumn{3}{c}{With STE} \\
         Method & AUPRC & AUP & AUR & AUPRC & AUP & AUR \\
         \midrule
         FreqShapes & 0.7789\std{0.0033} & 0.7324\std{0.0033} & 0.6561\std{0.0020} & \textbf{0.8905}\std{0.0018} & \textbf{0.7805}\std{0.0014} & \textbf{0.6618}\std{0.0019}\\ 
         SeqComb-MV & 0.7269\std{0.0033} & 0.8727\std{0.0009} & 0.3478\std{0.0022} & \textbf{0.7589}\std{0.0014} & \textbf{0.8783}\std{0.0007} & \textbf{0.3906}\std{0.0011}\\
         ECG & 0.6152\std{0.0007} & \textbf{0.7468}\std{0.0008} & 0.4023\std{0.0012} & \textbf{0.6599}\std{0.0009} & 0.7260\std{0.0010} & \textbf{0.4595}\std{0.0007}\\
         \bottomrule
    \end{tabular}
    \end{sc}
    }
    \end{center}
\end{table}

\begin{table}[t]
\scriptsize
    \centering
    \caption{Ablation of \modelname~considering whether there are different losses in our component.}
    \vspace{1mm}\label{tab:ablation_mbc_la}
    \resizebox{0.66\columnwidth}{!}{
    \begin{sc}
    \begin{tabular}{l|l|c c c}
         \toprule
         Dataset & Ablation & AUPRC & AUP & AUR\\
         \midrule
      & Full & \textbf{0.8905}\std{0.0018} & \textbf{0.7805}\std{0.0014} & {0.6618}\std{0.0019}\\
       FreqShapes  &  w/o  $\mathcal{L}_{\mathrm{LC}}$ & 0.2251\std{0.0014} & {0.1959}\std{0.0008} & \textbf{0.7675}\std{0.0015}\\
         &  w/o $\mathcal{L}_{\mathrm{KL}}$ & 0.8303\std{0.0027} &  0.6463\std{0.0027} &0.7034\std{0.0022}\\
          &  w/o $\mathcal{L}_{dr}$  &  0.1943\std{0.0017} &  0.1390\std{0.0015} & 0.4147\std{0.0019}\\
         \midrule
      & Full & \textbf{0.7589}\std{0.0014} & \textbf{0.8783}\std{0.0007} & {0.3906}\std{0.0011}\\
        SeqComb-MV  &  w/o  $\mathcal{L}_{\mathrm{LC}}$ & 0.0950\std{0.0033} & 0.0566\std{0.0018} & 0.3301\std{0.0080}\\
         &  w/o $\mathcal{L}_{\mathrm{KL}}$ & 0.7484\std{0.0016} &  0.8694\std{0.0008} & 0.3531\std{0.0013}\\
          &  w/o $\mathcal{L}_{dr}$  & 0.0688\std{0.0015} & 0.0532\std{0.0008} & \textbf{0.6279}\std{0.0075}\\
         \midrule
          & Full & \textbf{0.6599}\std{0.0009} & \textbf{0.7260}\std{0.0010} & \textbf{0.4595}\std{0.0007}\\
        ECG  &  w/o  $\mathcal{L}_{\mathrm{LC}}$ & {0.6209}\std{0.0019} & {0.6417}\std{0.0020} & 0.4287\std{0.0015}\\
         &  w/o $\mathcal{L}_{\mathrm{KL}}$ & 0.6417\std{0.0019} & 0.6979\std{0.0009} & 0.4424\std{0.0007}\\
          &  w/o $\mathcal{L}_{dr}$  & 0.1516\std{0.0003} &  0.1405\std{0.0003} & 0.6313\std{0.0006}\\
         \bottomrule
    \end{tabular}
    \end{sc}
    }
\end{table}

\paragraph{Effect of Different Losses.} We now examine the effectiveness of different losses in the model components.
We select the label consistency loss $\mathcal{L}_{\mathrm{LC}}$, the maintenance loss $\mathcal{L}_{\mathrm{KL}}$, and the loss of the reference distance $\mathcal{L}_{dr}$.
The results of our \modelname~with/without these losses are shown in Table~\ref{tab:ablation_mbc_la}, where containing all the losses of our method shows the best explanation performance.
Specifically, the absence of $\mathcal{L}_{\mathrm{KL}}$ drops some of the performance,  as the gap between the distribution of $X$ and the distribution of $\widetilde{X}$ becomes larger, and it is not enough to rely on $\widetilde{X}^r$ alone to do the perturbation.
Failure to predict an explanation when there is no $\mathcal{L}_{dr}$ is because there is not a reference explanation to generate, which is common sense.
The simulation datasets (FreqShape and SeqComb-MV) likewise failed to explain when there was no consistency labeling $\mathcal{L}_{\mathrm{LC}}$, while the ECG was able to be somewhat normal. It is because \modelname~generates $\widetilde{X}$ without label leakage while within the sample distribution.
This explains why the alone loss results observed individually perform poorly, while together these losses provide a powerful objective to obtain explanatory performance.

\begin{figure}[H]
	\centering
	\subfigbottomskip=0pt
	\subfigcapskip=-5pt 
		\subfigure[Freqshape]{\label{stab1}
			\includegraphics[width=.32\linewidth]{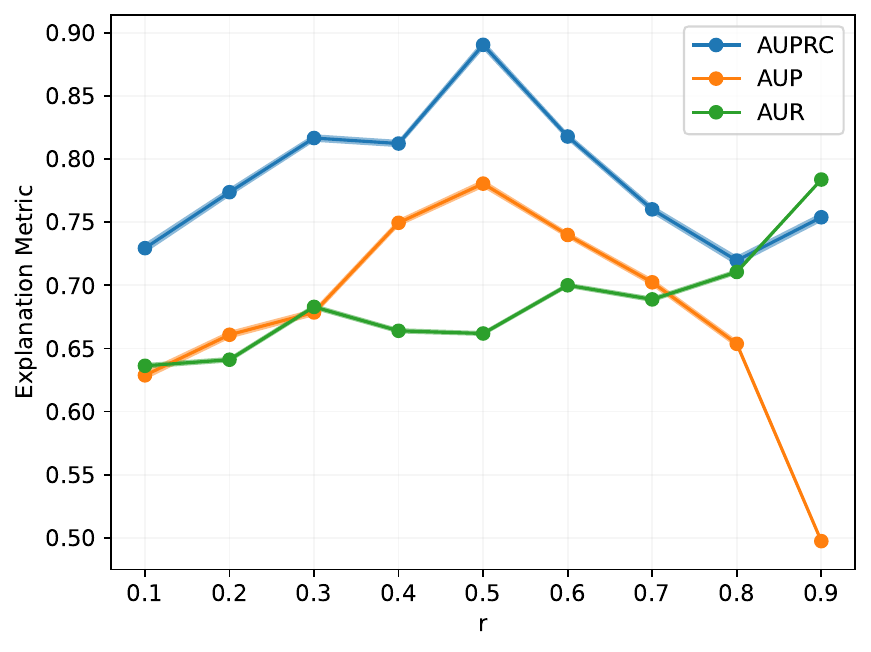}}
   	\subfigure[SeqComb-MV]{\label{stab2}
			\includegraphics[width=.32\linewidth]{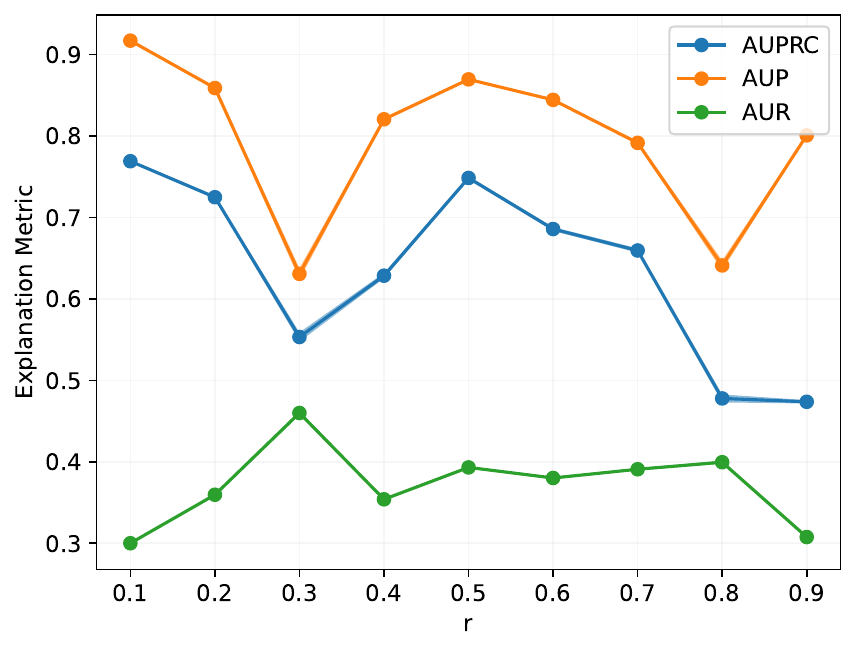}}
	\subfigure[ECG]{\label{stab3}
		\includegraphics[width=.32\linewidth]{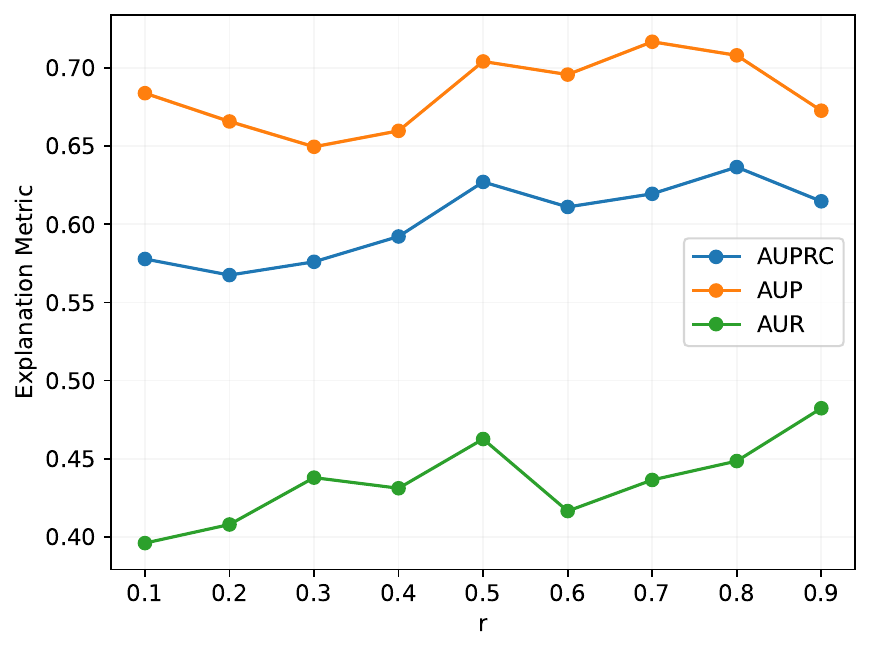}}\\
	\caption{Experiment on different datasets varying the $r$ parameter.
}\label{rvalues}
\end{figure}

\paragraph{Choosing the Parameter $r$.} One of the most significant parameters in training \modelname~is $r$, which governs the sparsity of the masks that are learned during the process. 
We conduct experiments on the above tree datasets to scrutinize the impact of varying $r$ on the quality of explanations generated by the model, where we hold all hyperparameters constant while varying $r$.
The outcomes of this variation are graphically represented in Figure~\ref{rvalues}, facilitating a visual interpretation of the relationship between the sparsity parameter $r$ and the explanation quality.
Lower values of the parameter are associated with a decrease in the performance of the explainer, as evidenced by the decline in AUR.
It is worth noting that the performance stabilizes when the value is between $0.4$ and $0.7$, suggesting that the effectiveness of the interpreter is relatively insensitive to the exact choice of $r$ in this range.
Consequently, a value proximate to $0.5$ is recommended for our experimental applications, a guideline that was adhered to in the course of this work.

\section{Different Black-box Classifications}
To explore the flexibility of \modelname, we study different time series classifiers and explore their explanatory role.
We replace the original transformer-based black-box $f$ to a long-short term memory~(LSTM) or a convolutional neural network~(CNN)  as the underlying classifiers with the following hyperparameters: (i) For LSTM, we use $3$ layer bidirectional LSTM and an MLP on the mean of last hidden states. (ii) For CNN, we use $3$ layer CNN and an MLP on meanpool.

We compare \modelname~against strong baselines: IG, Dynamask, WinIT, and \textsc{TimeX}. The results of the LSTM predictor are shown in Table~\ref{table:lstm_synth}-\ref{table:lstm_ecg}. 
\modelname~retains the best AUPRC prediction on those datasets and is also slightly ahead for AUP and AUR overall, while \textsc{TimeX} fails to predict on SeqComb-MV due to non-convergence.
Tables~\ref{table:cnn_synth}-\ref{table:cnn_ecg} show the results of our method against strong baselines with a CNN predictor.
Our method performs very well for both SeqComb-MV and ECG datasets, achieving the highest AUPRC and AUP for both datasets.
However, the performance for FreqShapes AUPRC has high values for both ours and IG, making the comparison more difficult.
Overall, our \modelname~maintains a relatively strong explanation performance among other black-box classifier architectures.

\begin{table}[H]
  \scriptsize
  \centering
  \caption{Explainer results with LSTM predictor on FreqShapes and SeqComb-MV synthetic datasets.}
  \vspace{1mm}
      \resizebox{0.9\columnwidth}{!}{
    \begin{sc}
  \begin{tabular}{l|c c c|c c c}
    \toprule
     & \multicolumn{3}{c|}{FreqShapes} & \multicolumn{3}{c}{SeqComb-MV}\\
     Method & AUPRC & AUP & AUR & AUPRC & AUP & AUR \\
    \midrule
     IG & 0.9282\std{0.0016} & 0.7775\std{0.0010} & 0.6926\std{0.0017} & 0.2369\std{0.0020} & 0.5150\std{0.0048} & 0.3211\std{0.0032} \\
     Dynamask & 0.2290\std{0.0012} & 0.3422\std{0.0037} & 0.5170\std{0.0013} & 0.2836\std{0.0021} & 0.6369\std{0.0047} & 0.1816\std{0.0015} \\
     WinIT & 0.4171\std{0.0016} & 0.5106\std{0.0026} & 0.3909\std{0.0017} & \underline{0.3515}\std{0.0014} & \underline{0.6547}\std{0.0026} & 0.3423\std{0.0021} \\
     TimeX & \underline{0.9903}\std{0.0002} & \textbf{0.7887}\std{0.0008} & \underline{0.7963}\std{0.0013} & 0.1298\std{0.0017} & 0.1307\std{0.0022} & \textbf{0.4751}\std{0.0015} \\
     \midrule
     \modelname & \textbf{0.9939}\std{0.0002} & \underline{0.7413}\std{0.0009} & \textbf{0.8428}\std{0.0008} & \textbf{0.4052}\std{0.0038} & \textbf{0.6804}\std{0.0052} & \underline{0.3519}\std{0.0021} \\
    \bottomrule
  \end{tabular}
    \end{sc}
  }
  \label{table:lstm_synth}
\end{table}

\begin{table}[H]
\scriptsize
\centering
\caption{Explainer results with LSTM predictor on ECG dataset.}
\vspace{1mm}
\resizebox{0.5\columnwidth}{!}{
    \begin{sc}
\begin{tabular}{l|c c c}
\toprule
 & \multicolumn{3}{c}{ECG}\\
 Method & AUPRC & AUP & AUR \\
\midrule
 IG & 0.5037\std{0.0018} & 0.6129\std{0.0026} & 0.4026\std{0.0015} \\
 Dynamask & 0.3730\std{0.0012} & 0.6299\std{0.0030} & 0.1102\std{0.0007} \\
 WinIT & 0.3628\std{0.0013} & 0.3805\std{0.0022} & 0.4055\std{0.0009} \\
 TimeX & \underline{0.6057}\std{0.0018} & \underline{0.6416}\std{0.0024} & \underline{0.4436}\std{0.0017} \\
 \midrule
\modelname & \textbf{0.6512}\std{0.0011} & \textbf{0.7432}\std{0.0011} & \textbf{0.4451}\std{0.0008} \\
\bottomrule
\end{tabular}
    \end{sc}
}
\label{table:lstm_ecg}
\end{table}

\begin{table}[H]
  \scriptsize
  \centering
    \caption{Explainer results with CNN predictor on FreqShapes and SeqComb-MV synthetic datasets.}
\vspace{1mm}
        \resizebox{0.9\columnwidth}{!}{
    \begin{sc}
  \begin{tabular}{l|c c c|c c c}
    \toprule
     & \multicolumn{3}{c|}{FreqShapes} & \multicolumn{3}{c}{SeqComb-MV}\\
     Method & AUPRC & AUP & AUR & AUPRC & AUP & AUR \\
    \midrule
     IG & \textbf{0.9905}\std{0.0007} & \textbf{0.8777}\std{0.0009} & 0.7056\std{0.0017} & 0.5979\std{0.0027} & \underline{0.8858}\std{0.0014} & 0.2294\std{0.0013}\\
     Dynamask & 0.2574\std{0.0008} & 0.4432\std{0.0032} & 0.5257\std{0.0015} & 0.4550\std{0.0016} & 0.7308\std{0.0025} & 0.3135\std{0.0019} \\
     WinIT &  0.5321\std{0.0018} & 0.6020\std{0.0025} & 0.3966\std{0.0017} & 0.5334\std{0.0011} & 0.8324\std{0.0020} & 0.2259\std{0.0020} \\
     TimeX & 0.7489\std{0.0046} & 0.4966\std{0.0033} & \underline{0.7916}\std{0.0021} & \underline{0.7016}\std{0.0019} & 0.7670\std{0.0012} & \textbf{0.4689}\std{0.0016} \\
     \midrule
     \modelname & \underline{0.9134}\std{0.0014} & \underline{0.6066}\std{0.0011} & \textbf{0.7952}\std{0.0014} & \textbf{0.7822}\std{0.0012} & \textbf{0.8896}\std{0.0005} & \underline{0.3434}\std{0.0012} \\
    \bottomrule
  \end{tabular}
    \end{sc}
  }
  \label{table:cnn_synth}
\end{table}

\begin{table}[H]
\scriptsize
\centering
\caption{Explainer results with CNN predictor on ECG dataset.}
\vspace{1mm}
\resizebox{0.5\columnwidth}{!}{
    \begin{sc}
\begin{tabular}{l|c c c}
\toprule
 & \multicolumn{3}{c}{ECG}\\
 Method & AUPRC & AUP & AUR \\
\midrule
 IG & 0.4949\std{0.0010} & 0.5374\std{0.0012} & \textbf{0.5306}\std{0.0010} \\
 Dynamask & 0.4598\std{0.0010} & 0.7216\std{0.0027} & 0.1314\std{0.0008} \\
 WinIT & 0.3963\std{0.0011} & 0.3292\std{0.0020} & 0.3518\std{0.0012} \\
 TimeX & \underline{0.6401}\std{0.0010} & \underline{0.7458}\std{0.0011} & 0.4161\std{0.0008} \\
 \midrule
\modelname & \textbf{0.6726}\std{0.0010} & \textbf{0.7570}\std{0.0011} & \underline{0.4319}\std{0.0012} \\
\bottomrule
\end{tabular}
    \end{sc}
}
\label{table:cnn_ecg}
\end{table}

\section{Visualization of FreqShape Dataset}\label{visme}
\begin{figure}[H]
    \centering
    \includegraphics[width=0.7\columnwidth]{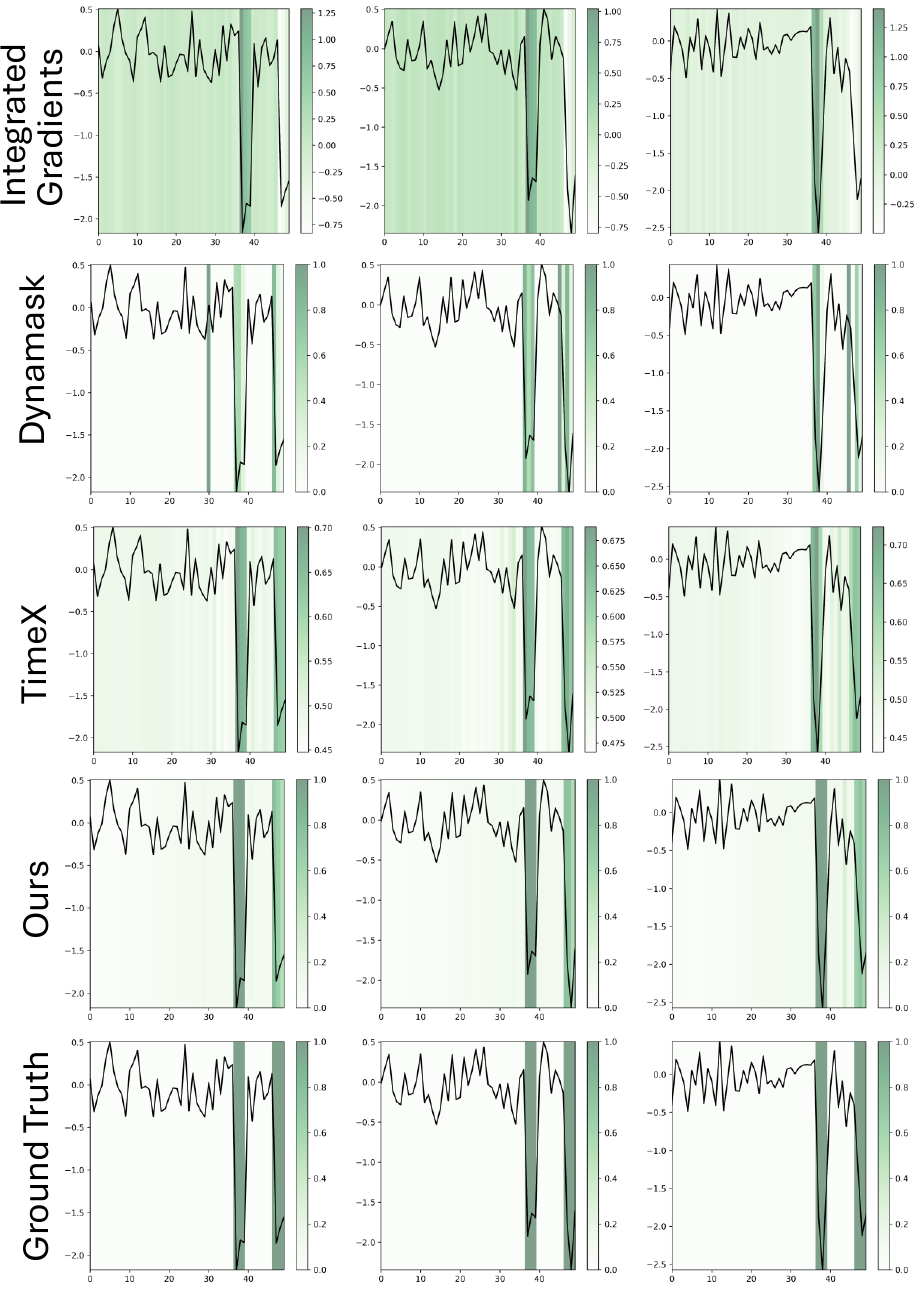}
    \vspace{-3mm}
    \caption{Visualization of all explainers on the FreqShapes dataset. Each column corresponds to a unique sample.  For each row, the method used to generate the corresponding explanation figure is indicated, with the ground truth explanations presented in the bottom row.}
    \vspace{-10mm}
    \label{fig:vis_methods}
\end{figure}

Saliency maps are indeed a potent tool for visualizing the significance of features, particularly in multivariate time series analysis~\cite{crabbe2021explaining, liu2024explaining}.
Thus, we demonstrate the saliency maps of the benchmarks and \modelname~on the FreqShapes dataset, where increasing and decreasing subsequences determine the class label.
Figure~\ref{fig:vis_methods} illustrates these explainers and the corresponding ground-truth in a sample.
IG identifies large areas as important and may prove to be untenable when applied to larger datasets replete with noise, where the pertinent signal may be less discernible. 
Dynamask seems to ignore several key sub-instances, often detecting only one or two salient segments with erroneous values.
\textsc{TimeX} is unable to describe important sub-instances with certainty, possibly due to constructed classifiers that produce data distribution bias.
In stark contrast, our method focuses on the peaks for matching to ground-truth explanations.

\end{document}